\documentclass{article}

\usepackage{comment}

    \usepackage[final,nonatbib]{neurips_2024}

\usepackage[utf8]{inputenc} % allow utf-8 input
\usepackage[T1]{fontenc}    % use 8-bit T1 fonts
\usepackage{hyperref}       % hyperlinks
\usepackage{url}            % simple URL typesetting
\usepackage{booktabs}       % professional-quality tables
\usepackage{amsfonts}       % blackboard math symbols
\usepackage{nicefrac}       % compact symbols for 1/2, etc.
\usepackage{microtype}      % microtypography
\usepackage{xcolor}         % colors
\usepackage{amsmath}
\usepackage{algorithm}
\usepackage{algpseudocode}
\usepackage{multirow}
\usepackage{multicol}
\usepackage{graphicx}
\usepackage{caption}
\usepackage{subcaption}
\usepackage{enumitem}
\usepackage{wrapfig}
\usepackage{amsthm}

\newcommand{\zz}[1]{{\color{red}[zz: #1]}}

\newtheorem{theorem}{Theorem}[section]

\newtheorem{proposition}[theorem]{Proposition}

 \usepackage{tikz}

\DeclareMathAlphabet\mathbfcal{OMS}{cmsy}{b}{n}
\newcommand{\ten}[1]{\mathbfcal{#1}} %mathcal
\newcommand{\tten}[1]{\mathbfcal{\widetilde{#1}}} %mathcal
\newcommand{\mat}[1]{\mathbf{#1}}
\newcommand{\tmat}[1]{\mathbf{\widetilde{#1}}}

\newcommand{\einsum}{\mbox{\bf einsum}}

\newcommand{\cI}{\mathcal{I}}

\title{CoMERA: Computing- and Memory-Efficient Training via Rank-Adaptive Tensor Optimization}

\author{%
Zi Yang \thanks{The majority of this work was done when the first author was a postdoc at UC Santa Barbara. }\\ 
University at Albany, SUNY \\
\texttt{zyang8@albany.edu} \\
\And 
Ziyue Liu\\
University of California at Santa Barbara \\
\texttt{ziyueliu@ucsb.edu}
\And 
Samridhi Choudhary \\
Amazon Alexa AI \\
\texttt{samridhc@amazon.com} \\
\And 
Xinfeng Xie \\
Meta \\
\texttt{xinfeng@meta.com} \\
\And
Cao Gao \\
Meta \\
\texttt{caogao@meta.com}\\
\And 
Siegfried Kunzmann \\
Amazon Alexa AI \\
\texttt{kunzman@amazon.com}\\
\And 
Zheng Zhang \\
University of California at Santa Barbara \\
\texttt{zhengzhang@ece.ucsb.edu}
}

\begin{document}

\maketitle

\begin{abstract}
Training large AI models such as LLMs and DLRMs costs massive GPUs and computing time. The high training cost has become only affordable to big tech companies, meanwhile also causing increasing concerns about the environmental impact. This paper presents {CoMERA}, a \underline{\bf Co}mputing- and \underline{\bf M}emory-\underline{\bf E}fficient training method via \underline{\bf R}ank-\underline{\bf A}daptive tensor optimization. CoMERA achieves rank-adaptive tensor-compressed (pre)-training via a multi-objective optimization formulation and improves the training to provide both a high compression ratio and excellent accuracy in the training process. Our optimized numerical computation (e.g., optimized tensorized embedding and tensor-network contractions) and GPU implementation eliminate part of the run-time overhead in the tensorized training on GPU. This leads to, for the first time, $2-3\times$ speedup per training epoch compared with standard training. CoMERA also outperforms the recent GaLore in terms of both memory and computing efficiency. Specifically, CoMERA is $2\times$ faster per training epoch and $9\times$ more memory-efficient than GaLore on a tested six-encoder transformer with single-batch training.  Our method also shows $\sim 2\times$ speedup than standard pre-training on a BERT-like code-generation LLM while achieving $4.23\times$  compression ratio in pre-training.
With further HPC optimization, CoMERA may reduce the pre-training cost of many other LLMs. 
 An implementation of CoMERA is available at \url{https://github.com/ziyangjoy/CoMERA}. 
%We further show that the efficiency can be further boosted by employing low-/mixed-precision low-rank tensor computation. 
\end{abstract}

\section{Introduction} \label{sc:intro}
Deep neural networks have gained success in solving numerous engineering problems. These approaches usually use a huge number of variables to parametrize a network, and require massive hardware resources to train the model. For instance, the Deep Learning Recommendation Model (DLRM) released by Meta (which is smaller than the practical product)~\cite{naumov2019deep} has 4.2 billion parameters; GPT-3~\cite{brown2020language} has 175 billion parameters. %Due to the huge model size, high-performance computing (HPC) platforms are usually required to handle the expensive training. 
OpenAI shows that the computing power required for key AI tasks has doubled every 3.4 months~\cite{OpenAI} since 2012. Training a large language model like ChatGPT and LLaMA from scratch often takes several weeks or months on thousands of GPUs~\cite{chatGPT-cost,CNBC}. 

% Building ever-larger AI models entails a tremendous amount of energy expenditure and thus carbon emissions, leading to an energy consumption growing at a breathtaking rate. The study~\cite{strubell2019energy} estimated that training a transformer for NLP can generate up to 626,155 pounds of CO$_2$ emissions—roughly equal to the total lifetime carbon footprint of five cars. As a comparison, the average American generates 36,156 pounds of CO$_2$ emissions in a year. The study in~\cite{patterson2021carbon} further shows that training GPT-3 takes 1.287 gigawatt hours (or about as much electricity as 120 US homes would consume in a year) and generates 502 tons of carbon emissions (or about as much as 110 US cars emit in a year in the research). Pre-traning a large AI model normally requires many training runs, indicating that the total energy cost can be even tremendously higher.

Large AI models often have much redundancy. Therefore, numerous methods have been developed to reduce the cost of AI inference~\cite{calvi2019compression,kim2015compression,chen2015compressing,han2015deep,he2016effective,molchanov2019importance,luo2017thinet,hinton2015distilling}). However, training large AI models (especially from scratch) remains an extremely challenging task. Low-precision training~\cite{hubara2017quantized,koster2017flexpoint,gupta2015deep,sun2020ultra} has been popular in on-device setting, but its memory reduction is quite limited. Furthermore, it is hard to utilize ultra low-precision training on GPU since current GPUs only support limited precision formats for truly quantized training. Chen~\cite{chen2021pixelated} employed the idea of robust matrix factorization to reduce the training cost. Similar low-rank matrix approximation techniques have been applied to train large AI models including large language models~\cite{zhao2024galore,huh2024training}. Among them, GaLore~\cite{zhao2024galore} can train the 7B LLaMA model on an RTX 4090 GPU using a single-batch and layer-wise setting. However, this setting can lead to extremely long training time, which is infeasible in practical settings. 

% Actually, the pre-training cost of state-of-the-art AI models (e.g., language/vision foundation models) are so high that only a small number of big tech companies can afford the cost. 
% While pruning techniques~\cite{neklyudov2017structured,wen2016learning} can generate memory-efficient models for inference, they do not always reduce the memory and energy cost of pre-training because the sparsity pattern is unknown {\it a-priori}. 

Compared with matrix compression, low-rank tensor compression has achieved much higher compression ratios on various neural networks~\cite{gu2022heat,jaderberg2014speeding,kim2015compression,lebedev2014speeding,novikov2015tensorizing,japanttrnn,ttrnn2017icml,zheng2024svdinstn}. This idea has been studied in structure search for compact representations~\cite{zheng2024svdinstn}, post-training compression~\cite{gu2022heat}, fixed-rank training ~\cite{calvi2019compression,novikov2015tensorizing,khrulkov2019tensorized}, zeroth-order training~\cite{zhao2023tensor} and parameter-efficient fine tuning~\cite{yang2024loretta}. The recent work~\cite{hawkins2022towards,hawkins2021bayesian} provides a rank-adaptive tensor-compressed training from a Bayesian perspective. However, to achieve a reduction in both memory and training time (especially on transformers), two open problems need to be addressed. Firstly, a more robust rank-adaptive tensor-compressed training model is desired, since the method in~\cite{hawkins2022towards,hawkins2021bayesian} relies on a heuristic fixed-rank warm-up training. Secondly, while modern GPUs are well-optimized for large-size matrix computations, they are unfriendly for low-rank tensor-compressed training. Specifically, most operations in tensor-compressed training are small-size tensor contractions, which can cause significant runtime overhead on GPUs even though the theoretical computing FLOPS is very low. As a result, as far as we know no papers have reported real training speedup on GPU. This issue was also observed in~\cite{hrinchuk2019tensorized}. SVDinsTN~\cite{zheng2024svdinstn} controls tensor ranks to search for a compact tensor structure of a given tensor. HEAT~\cite{gu2022heat} uses tensor decompositions for post-training model compression of trained models. Detailed comparisons with these works are shown in Appendix \ref{appendix:comparison_existing}.

\begin{figure}[t]
     \centering
     \begin{subfigure}[b]{0.48\textwidth}
         \centering
         \includegraphics[width=\textwidth]{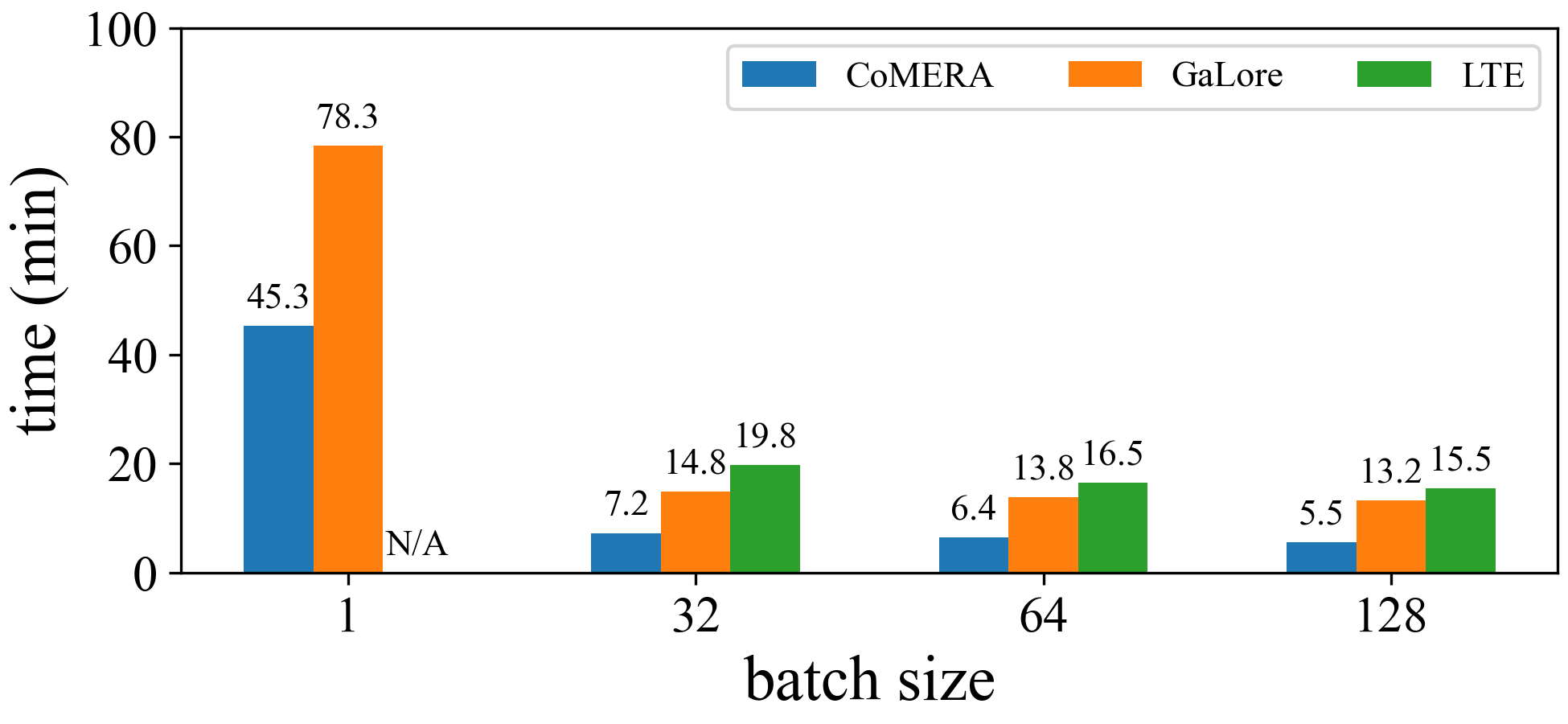}
         \caption{Training time per epoch.}
     \end{subfigure}
     \hfill
     \begin{subfigure}[b]{0.48\textwidth}
         \centering
         \includegraphics[width=\textwidth]{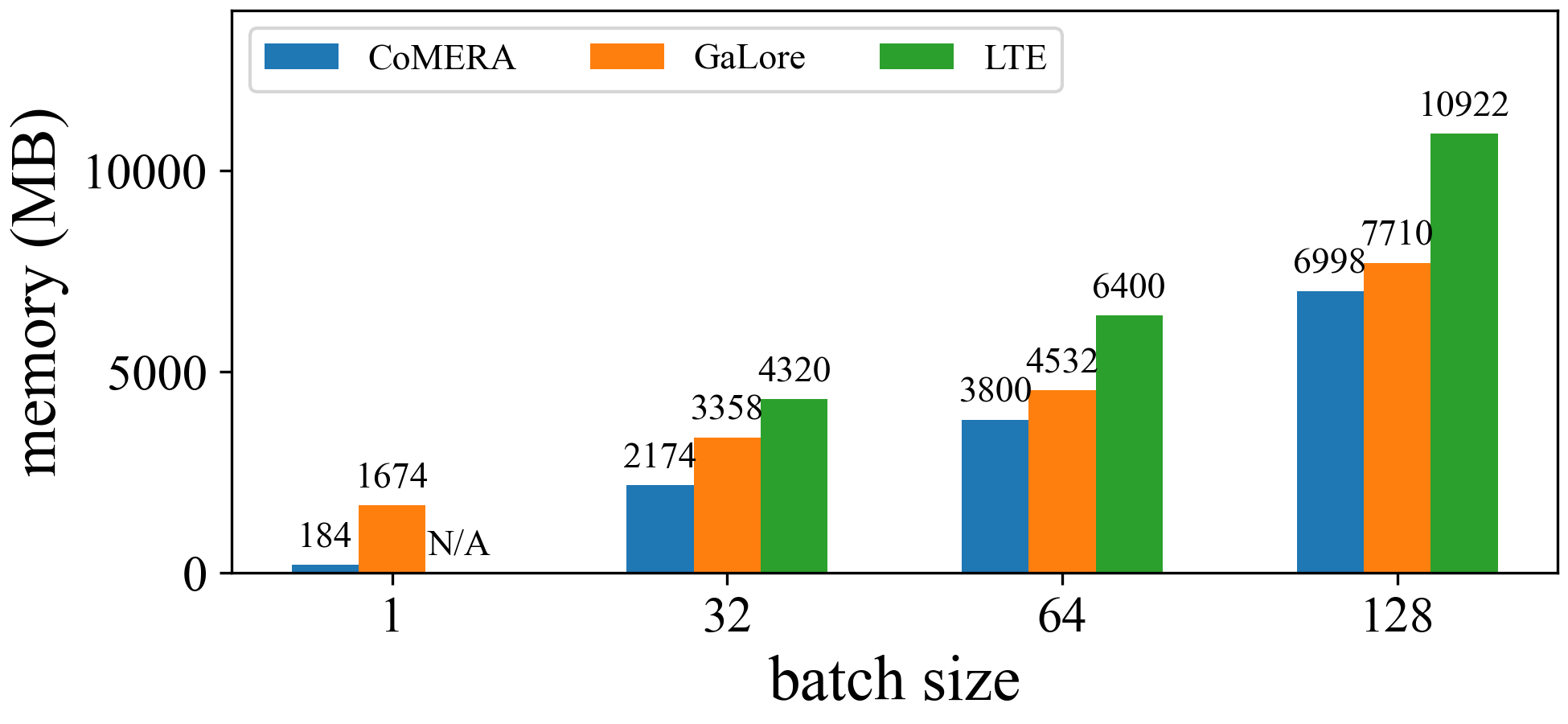}
         \caption{Peak memory consumption.}
     \end{subfigure}
     \caption{Training time and total memory cost of CoMERA, GaLore~\cite{zhao2024galore} and LTE~\cite{huh2024training} on a six-encoder transformer with varying batch sizes. The experiment is done on Nvidia RTX 3090 GPU.}
     \label{fig:compare_galore_LTE}
     \vspace{-15pt}
\end{figure}

\vspace{-10pt}
\paragraph{Paper Contributions.} In this work, we propose CoMERA, a tensor-compressed training method that can achieve, for the first time,  {\it simultaneous reduction of both memory and runtime on GPU}. Our specific contributions are summarized as follows.
\vspace{-5pt}
\begin{itemize}[leftmargin=*]
\vspace{-4pt}
    \item {\bf Multi-Objective Optimization for Rank-Adaptive Tensor-Compressed Training}. We propose a multi-objective optimization formulation to balance the compression ratio and model accuracy and to customize the model for a specific resource requirement. One by-product of this method is the partial capability of automatic architecture search: some layers are identified as unnecessary and can be completely removed by rank-adaptive training. 

    % \item We propose the gradient normalization method to stabilize the optimization for tensor-compressed training. It normalizes the update of each tensor core based on its norm to ensure all tensor cores are equally contributing to the layer. This method is an add-on to all existing optimization algorithms and makes tensor-compressed optimization more stable and robust. 
\vspace{-4pt}
    \item {\bf Performance Optimization of Tensor-Compressed Training.} While tensor-compressed training greatly reduces the memory cost and computing FLOPS, it often slows down the practical training on GPU. We propose three approaches to achieve real training speedup: \textcircled{1} optimizing the lookup process of tensorized embedding tables, \textcircled{2} optimizing the tensor-network contractions in both forward and backward propagation, \textcircled{3} eliminating the GPU backend overhead via CUDA Graph.  %We propose a low-precision training framework for tensor-compressed training. A naive adoption of low-precision to tensor-compressed models may even slow down the training. We analyze the reason causing the slow-down and provide a mixed-precision training algorithm to maximize the benefits of low-precision computation in tensor-compressed models. 
\vspace{-4pt}
    \item {\bf Experimental Results.} We evaluate our method on the end-to-end training of a transformer with six encoders and the deep learning recommendation system model (DLRM). On these two benchmarks, our method achieves $80\times$ and $99\times $ compression ratios respectively, while maintaining the testing accuracy of standard uncompressed training. CoMERA also achieves $2-3\times$ speedup per training epoch compared with standard training methods on the transformer model. In a preliminary study of LLM pre-training, CoMERA shows $1.9\times$ to $2.3\times$ speedup in different pre-training stages on CodeBERT~\cite{husain2019codesearchnet}, while achieving $4.23\times$ overall model reduction in the pre-training process.
\end{itemize}

Figure \ref{fig:compare_galore_LTE} compares our CoMERA with GaLore~\cite{zhao2024galore} and the recent LoRA-based training method LTE~\cite{huh2024training} on the six-encoder transformer. When data and back-end memory cost are considered, CoMERA's memory consumption is $9\times$ less than GaLore in the single-batch training as adopted in~\cite{zhao2024galore}, and it uses the least memory under all batch sizes. Our method is $2-3\times$ faster than GaLore and LTE in each training epoch, although CoMERA has not yet been fully optimized on GPU.

While this work focuses on reducing the memory and computing cost of training, it can also reduce the communication cost by orders of magnitude: only low-rank tensorized model parameters and gradients need to be communicated in a distributed setting. The CoMERA framework can also be implemented on resource-constraint edge devices to achieve energy-efficient on-device learning. 

\vspace{-10pt}
\section{Background}
\vspace{-5pt}
\label{sc:preliminary}
% {\bf Notation.} Throughout the paper, lower-case letters (e.g., $a$) denote scalars; lower-case bold letters (e.g., $\mat{a}$) denote vectors; upper-case bold letters (e.g., $\mat{A}$) denote matrices; upper-case calligraphic bold letters (e.g., $\ten{A}$) denote tensors. $\mat{1}$ or $\mat{0}$ are a vector/matrix whose entries are all $1$ or $0$, respectively. $\mat{I}_n$ is an $n$-by-$n$ identity matrix. We use $[n]$ to denote the set of integers $\{1,2,\ldots,n\}$.  For a vector $\mat{v}$, $\|\mat{v}\|$ and $\|\mat{v}\|_1$denote its Euclidean norm and $1$-norm, respectively. $\|\mat{v}\|_0$ is the $0$-norm, that is the number of nonzero entries in $\|\mat{v}\|_0$. For a matrix $\mat{A}$, $\mat{A}^T$ denotes its transpose; $\|\mat{A}\|$ represents the Frobenius norm, and the spectrum norm $\|\mat{A}\|_2$ is the largest singular value of $\mat{A}$. We use MATLAB-style indexing to denote submatrices. For instance, $\mat{A}(i_1:i_2,j_1:j_2)$ denotes the submatrix consisting of the rows from $i_1$ to $ i_2$ and the columns from $j_1$ to $j_2$.

%\subsection{Tensor and Tensor Contraction}
\begin{figure}[t]
    \centering
    \vspace{-15pt}
    \includegraphics[width=0.8\textwidth]{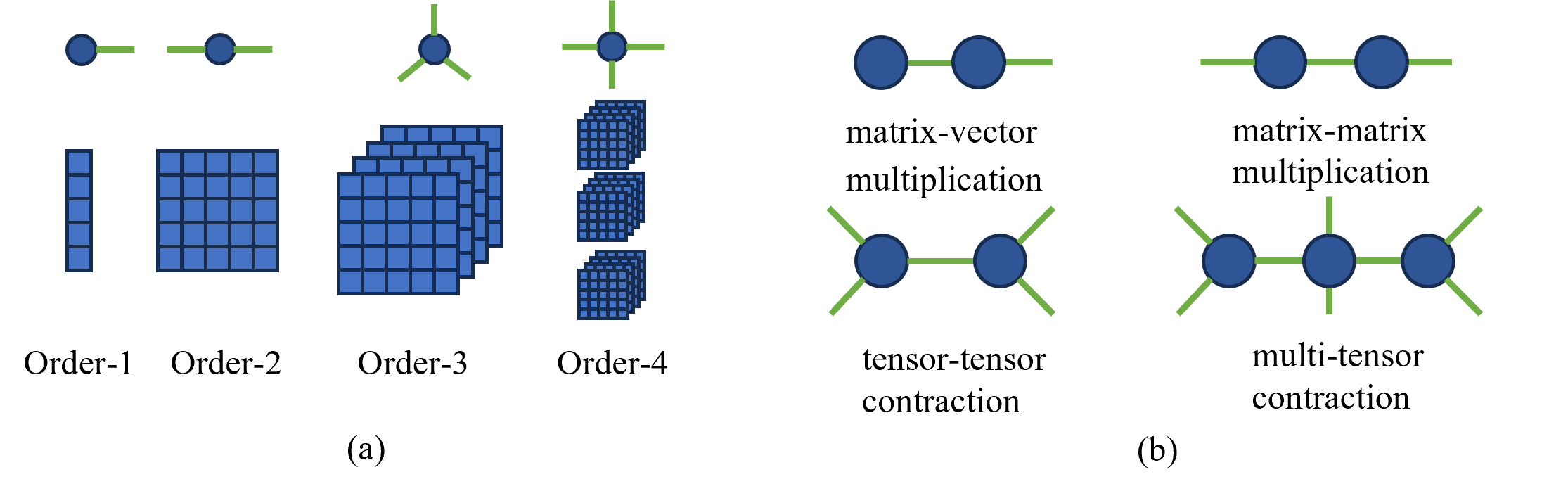}
    \caption{(a) Tensors. (b) Tensor contractions.}
    \label{fig:tensor}
    \vspace{-5pt}
\end{figure}

The tensor \cite{tensor:suvey,landsberg2012tensors} $\ten{A}\in \mathbb{R}^{n_1\times n_2 \times \cdots \times n_d}$ is indexed as $
    \ten{A} = (a_{i_1\cdots i_d})_{1\le i_j\le n_j}$ and is said to have order $d$ and dimension $n_1,
\ldots,n_d$. The Frobenius norm of tensor $\ten{A}$ is defined as $
    \|\ten{A} \|:= \sqrt{\sum_{i_1,\ldots,i_d}^{n_1,\ldots,n_d}a_{i_1\cdots i_d}^2}$. In tensor networks, the order-$d$ tensor $\ten{A}$ is represented as a node with $d$ edges. Some tensor network representations are illustrated in Fig. \ref{fig:tensor} (a). 
\vspace{-10pt}
\paragraph{Tensor Contraction.} Let $\ten{A} \in \mathbb{R}^{n_1\times n_2 \times \cdots \times n_d}$ and $\ten{B} \in \mathbb{R}^{l_1\times l_2 \times \cdots \times l_m}$ be two tensors with $n_s=l_t$. The tensor contraction product $\ten{C}=\ten{A}\times_{s,t} \ten{B}$ has dimension $\Pi_{i\neq s}n_i \times \Pi_{j\neq t}l_j$ and the entries are 
\begin{equation} \label{eq:contraction}
     c_{(i_p)_{p\neq s},(j_p)_{p\neq t}} = \sum_{i_s=j_t=1}^{n_s} a_{i_1\cdots i_s \cdots i_m} b_{j_1\cdots j_t \cdots j_k}.
\end{equation}
This definition can be naturally generalized to multiple pairs. Figure~\ref{fig:tensor}(b) illustrates some tensor contractions. For general operations among multiple tensors, we use PyTorch einsum in the following
\begin{equation}\label{eq:einsum}
    \ten{B} = \einsum(S_1,\ldots,S_m \Rightarrow T,[\ten{A}_1,\ldots,\ten{A}_m]),
\end{equation}
where each $S_i$ is a string of characters that specifies the dimension of $\ten{A}_i$. The output tensor $\ten{B}$ is obtained by summing over all other dimensions that are not in $T$. In the following, we show a few commonly used einsum operations. The Tensor-Train decomposition as in \eqref{eq:TT} is 
    \[
        \ten{A} = \einsum(n_1r_1,\ldots,r_{d-1}n_d \Rightarrow n_1n_2\cdots n_d,[\ten{G}_1,\ten{G}_2\ldots,\ten{G}_d]).
    \]
For the batched matrices $\ten{A}\in \mathbb{R}^{b\times m\times k},\ten{B}\in \mathbb{R}^{b\times k\times n}$, the batched matrix multiplication is 
    \[
        \ten{C} = \ten{A}\ten{B} = \einsum(bmk,bkn \Rightarrow bmn,[\ten{A},\ten{B}]),\; \; \text{where} \; \ten{C}[i,:,:] = \ten{A}[i,:,:]\ten{B}[i,:,:].
    \]

% The outer product of vectors $\{\mat{u}_i \in \mathbb{R}^{n_i}\}_{i=1}^d$ is the tensor
% \[
% \mat{u}_1\circ \mat{u}_2 \circ \cdots \circ \mat{u}_d := (\prod \limits_{k=1}^d \mat{u}_j [i_j])_{1\le i_j\le n_j} \in \mathbb{R}^{n_1\times n_2 \times \cdots \times n_d}.
% \]
% The tensors of the above form are called rank-1 tensors. Every tensor can be written as a sum of rank-1 tensors, i.e.,
% \[
%     \ten{A} = \sum_{i=1}^r \mat{u}_1^i\circ \cdots \circ \mat{u}_d^i.
% \]
% The above decomposition is called CP tensor decomposition \cite{tensor:suvey} and the smallest such integer $r$ is called the CP rank \cite{tensor:suvey} of $\ten{A}$, denoted by $\rank(\ten{A})$. The CP decomposition is illustrated in Figure \ref{fig:CP}.

%\subsection{Tensor Decomposition} 

\begin{figure}[h]
    \centering
    \vspace{-5pt}
    \includegraphics[width=0.8\textwidth,height=1in]{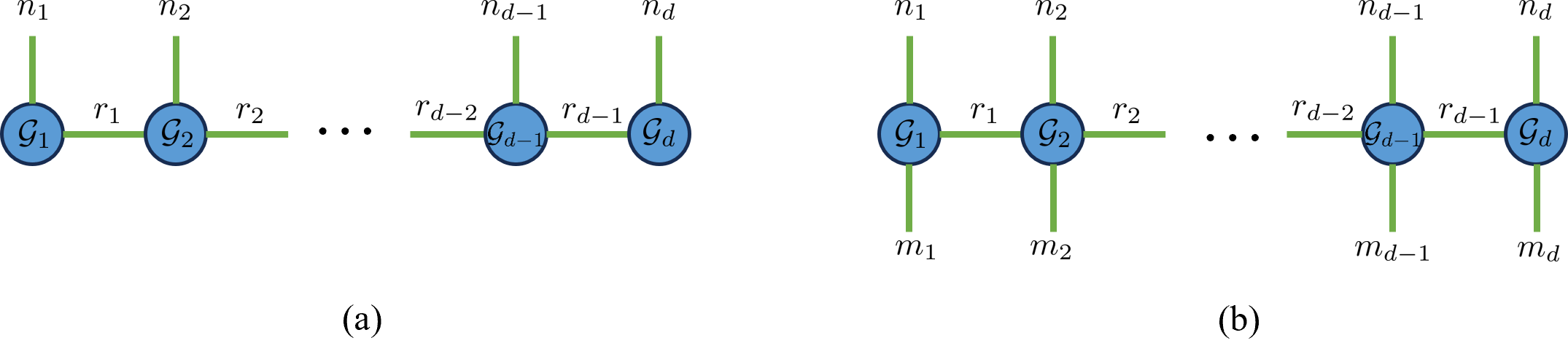}
    \caption{Tensor networks for (a) tensor-train and (b) tensor-train-matrix decompositions.}
    \label{fig:TT_network}
    \vspace{-10pt}
\end{figure}

\vspace{-10pt}
\paragraph{Tensor Decomposition.} In this paper, we will mainly use tensor-train (TT) \cite{oseledets2011tensor} and tensor-train matrix (TTM) \cite{novikov2015} decomposition for compressed neural network training. TT \cite{oseledets2011tensor} represents the tensor $\ten{A} \in \mathbb{R}^{n_1\times n_2 \times \cdots \times n_d}$ as a set of small-size cores 
$\ten{G}_1,\ldots,\ten{G}_{d}$ such that $\ten{G}_i \in \mathbb{R}^{r_{i-1}\times n_i \times r_{i}}$ and
\begin{equation} \label{eq:TT}
     \ten{A} = \ten{G}_1 \times_{3,1} \ten{G}_2 \times_{3,1} \cdots \times_{3,1} \ten{G}_d.
\end{equation}
The tuple $(r_0,r_1,\ldots,r_d)$ is the TT rank of the TT decomposition \eqref{eq:TT} and must satisfy $r_0=r_d=1$. 
%Figure \ref{fig:TT} shows the TT decomposition of an order-3 tensor and 
% Figure \ref{fig:TT_network}(a) shows the TT decomposition in the form of a tensor network.
% \begin{wrapfigure}{r}{2.6in}
%          \centering
%          \includegraphics[width=2.6in]{figures/TT.png}
%          \caption{Tensor-Train decomposition.}
%          \label{fig:TT}
% \end{wrapfigure}
TTM considers an order-$2d$ tensor $\ten{B}$ of dimension $ m_1 \times n_2 \times \cdots \times m_d \times n_d$, and represents $\ten{B}$ as 
\begin{equation}\label{eq:TTM}
    \ten{B} = \ten{F}_1 \times_{4,1} \ten{F}_2 \times_{4,1} \cdots \times_{4,1} \ten{F}_d,
\end{equation}
where $\ten{F}_i \in \mathbb{R}^{r_{i-1}\times m_i \times n_i \times r_{i}}$ for $i=1,\ldots,d$ and $r_0=r_d=1$. 
Figure~\ref{fig:TT_network} shows the tensor-network representations of TT and TTM decomposition.

% \begin{figure}[h]
%      \centering
%      \begin{subfigure}[b]{0.44\textwidth}
%          \centering
%          \includegraphics[width=\textwidth]{figures/TT_TTM_contractions.png}
%          \caption{Tensor-Train decomposition in tensor network representation}
%          \label{fig:TT_network}
%      \end{subfigure}
%      \hfill
%      \begin{subfigure}[b]{0.44\textwidth}
%          \centering
%          \includegraphics[width=\textwidth]{figures/TTM_network.png}
%          \caption{Tensor-Train-Matrix decomposition in tensor network representation}
%          \label{fig:TTM_network}
%      \end{subfigure}
% \end{figure}

In tensor-compressed neural networks, large weight matrices are reshaped to high-order tensors and compressed into small tensor cores in TT or TTM format. The weights of linear layers are often compressed into the TT format due to its efficiency in tensor-vector multiplications. The TTM format is more suitable for embedding tables whose dimension is highly unbalanced.

\vspace{-10pt}
\section{The CoMERA Training Framework}
\vspace{-5pt}
\label{sc:training}
%\subsection{Rank-Adaptive Tensor-Compressed Training as Multi-objective Optimization}
The size of the tensor-compressed neural networks can be adjusted by modifying the tensor ranks. However, it also brings in an important problem: \textit{how can we determine the tensor ranks automatically for a given resource limit?} We propose a multi-objective optimization to address this issue. 

\subsection{Multi-Objective Training Model}
\vspace{-5pt}
\paragraph{A Modified TT Representation.} We consider the tensor-compressed training for a generic neural network. Suppose that the neural network is parameterized as $f(\mat{x}|\{\ten{G}^i_{1},\ldots,\ten{G}^i_{d_i}\}_{i=1}^P )$, where $\{\ten{G}^i_{j} \in \mathbb{R}^{r^i_{j-1}\times n^i_{j} \times r^i_j}\}_{j=1}^{d_i}$ compress the original weight $\mat{W}_i$. Let $\{\mat{x}_k,y_k\}_{k=1}^N$ be training data and $\mathcal{L}$ be the loss function. The training is to minimize the following objective function 
\begin{equation} \label{eq:loss}
    \min_{\{\ten{G}^i_{1},\ldots,\ten{G}^i_{d_i}\}_{i=1}^P} L:=\sum_{k=1}^N \mathcal{L}(y_k,f(\mat{x}_k|\{\ten{G}^i_{1},\ldots,\ten{G}^i_{d_i}\}_{i=1}^P )).
\end{equation}
We modify the TT compression and control the ranks of $\ten{G}^i_{1},\ldots,\ten{G}^i_{d_i}$ by a set of diagonal matrices $\{\mat{D}^i_j \in \mathbb{R}^{r^i_j\times r^i_j} \}_{j=1}^{d-1}$. Specifically, let $\ten{W}_i$ be the reshape of $\mat{W}_i$, and the compression of $\mat{W}_i$ is 
\begin{equation} \label{eq:Wcompress}
    \ten{W}_i = \ten{G}^i_{1} \times_{3,1} \mat{D}^i_{1} \times_{2,1} \ten{G}^i_{2} \times_{3,1} \cdots \times_{3,1} \mat{D}^i_{d_i-1} \times_{2,1} \ten{G}^i_{d_i}.
\end{equation}
Now the tensor cores for $\ten{W}_i$ have 
$
    S_i = n_1^i\|\mat{D}_{1}^i\|_0 + n_{d_i}^i\|\mat{D}_{d_i-1}^i\|_0 + \sum_{j=2}^{d_i-1} n^i_j \|\mat{D}_{j-1}^i\|_0\|\mat{D}_{j}^i\|_0 
$ variables. 
For simplicity, we denote $\ten{G}:=\{\ten{G}^i_{1},\ldots,\ten{G}^i_{d_i}\}_{i=1}^P$ and $\mat{D}:= \{\mat{D}_1^i,\ldots,\mat{D}_{d_i-1}^i\}_{i=1}^P $.

\vspace{-5pt}
\paragraph{Multi-Objective Optimization.} We intend to minimize both the loss and compressed network size, which can be formulated as a multi-objective optimization 
$
    \min_{\ten{G},\mat{D}} \left\{ L(\ten{G},\mat{D}), S(\mat{D})\right\},
$
where $S(\mat{D}):=\sum_{i=1}^P S_i(\mat{D})$.
In most cases, we cannot find a point that minimizes the loss and model size simultaneously. Therefore, we look for a Pareto point $(\ten{G}^*,\mat{D}^*)$, meaning that there exist no $\ten{G}$ and $\mat{D}$ such that $L(\ten{G},\mat{D})\le L(\ten{G}^*,\mat{D}^*)$, $S(\mat{D}) \le S(\mat{D}^*)$, and at least one of inequalities is strict.

\vspace{-5pt}
\subsection{Training Methods}
\label{sc:training methods}
\vspace{-5pt}

We convert a multi-objective optimization to a single-objective one via scalarization. We use different scalarization methods at the early and late stage of training. The late stage is optional, and it can further compress the model to enable efficient deployment on resource-constraint platforms.
\vspace{-10pt}
\paragraph{Early Stage.} At the early stage of CoMERA, aggressively pruning ranks dramatically hurts the convergence. Hence, we start the training with the following linear scalarization formulation \cite{deb2016multi}
\begin{equation} \label{eq:pretrain-rank-origin}
    \min_{\ten{G},\mat{D}} \,\,  L(\ten{G},\mat{D})+\gamma {S}(\mat{D}).
\end{equation}
It is still hard to solve \eqref{eq:pretrain-rank-origin} since $S(\mat{D})$ uses $\|\cdot \|_0$ which is nonsmooth. Therefore, we replace $\|\cdot\|_0$ by the $\ell_1$ norm $\|\cdot\|_1$ and get the convex relaxation 
\begin{equation}
    \hat{S}(\mat{D}):= \sum_{i=1}^P \left(\sum_{i=1}^P n_1^i\|\mat{D}_{1}^i\|_1 + n_{d_i}^i\|\mat{D}_{d_i-1}^i\|_1 + \sum_{j=2}^{d_i-1} n^i_j \|\mat{D}_{j-1}^i\|_1\|\mat{D}_{j}^i\|_1 \right).
\end{equation}
We note that $\hat{S}(\mat{D})$ can be arbitrarily close to $0$ while keeping $L(\ten{G},\mat{D})$ unchangeable, since the corresponding slices of TT factors can be scaled accordingly. Therefore, a direct relaxation of the scalarization \eqref{eq:scalarization} does not have a minimizer. To address this issue, we add an $\ell_2$ regularization $\|\ten{G}\|^2:=\sum_{i=1}^P \sum_{j=1}^{d_i} \|\ten{G}^i_{j}\|^2$ to the relaxation and get the formulation
\begin{equation} \label{eq:pretrain-rank}
    \min_{\ten{G},\mat{D}} \,\,  L(\ten{G},\mat{D})+\gamma \hat{S}(\mat{D})+\beta \|\ten{G}\|^2.
\end{equation}
%Mathematically, the $l_2$ regularization term in \eqref{eq:pretrain-rank} is the same as bounding the Frobenius norm of tensor cores. 
The optimizer of Problem \eqref{eq:pretrain-rank} is a Pareto point for a constrained problem, shown in the following.

\begin{proposition}\label{prop:Pareto}
    For all $\gamma>0,\beta>0$, there exists some constant $C>0$ such that the solution to the problem \eqref{eq:pretrain-rank} is a Pareto point of the following multi-objective optimization problem
    \begin{eqnarray} \label{eq:constrained-pretrain-rank}
        \min_{\ten{G},\mat{D}} & (L(\ten{G},\mat{D}),\hat{S}(\mat{D})) & 
        \text{subject to}\, \|\ten{G}\|^2 \le C.
    \end{eqnarray}
\end{proposition}
\vspace{-10pt}
\begin{proof}
    See Appendix \ref{sec:proof:prop_Pareto} for the complete proof.
\end{proof}
\vspace{-10pt}

\vspace{-10pt}
\paragraph{Late Stage (Optional).} The early-stage training can provide us with a Pareto point, but we cannot control where the Pareto point is. In the late stage of CoMERA, we may continue training the model towards a preferred loss $L_0$ and a preferred model size $S_0$ for deployment requirements. This can be achieved by the achievement scalarization \cite{deb2016multi} that leads to a Pareto point close to $(L_0,S_0)$: 
\begin{equation} \label{eq:scalarization}
    \min_{\ten{G},\mat{D}} \,\, \max \left\{ w_1 (L(\ten{G},\mat{D})-L_0), w_2 (S(\mat{D}) - S_0)\right\} + \rho (L(\ten{G},\mat{D})+S(\mat{D})).
\end{equation}
Here $w_1,w_2>0$ scale the objectives into proper ranges, and $\rho>0$ is a small constant. After relaxing $S(\mat{D})$ to $\hat{S}(\mat{D})$ and adding the regularization term, we get the following problem 
\begin{equation} \label{eq:scalarization relax}
    \min_{\ten{G},\mat{D}} \,\, \max \left\{ w_1 (L(\ten{G},\mat{D})-L_0), w_2 ({S}(\mat{D}) - S_0)\right\} + \rho (L(\ten{G},\mat{D})+\hat{S}(\mat{D}))+\beta \|\ten{G}\|^2,
\end{equation}
where $\beta>0$ is a positive constant. Note that the ${S}(\mat{D})$ inside $\max$ is not relaxed now for accurate comparisons. When $w_1 (L(\ten{G},\mat{D})-L_0)\ge w_2 ({S}(\mat{D}) - S_0)$, we consider the following problem 
\begin{equation}\label{eq:1>2}
    \min_{\ten{G},\mat{D}} \,\, w_1 (L(\ten{G},\mat{D})-L_0) + \rho (L(\ten{G},\mat{D})+\hat{S}(\mat{D}))+\beta \|\ten{G}\|^2.
\end{equation}
We run a step of a gradient-based algorithm on this problem. 
When $w_1 (L(\ten{G},\mat{D})-L_0)< w_2 ({S}(\mat{D}) - S_0)$, we relax the $S(\mat{D})$ again and get the following problem 
\begin{equation} \label{eq:1<2}
    \min_{\ten{G},\mat{D}} \,\, w_2 (\hat{S}(\mat{D}) - S_0) + \rho (L(\ten{G},\mat{D})+\hat{S}(\mat{D}))+\beta \|\ten{G}\|^2,
\end{equation}
and run a step of a gradient-based algorithm on this problem. The Algorithm \ref{alg:scalar} is summarized in Appendix \ref{appendix:late stage}. The late stage optimization can be independently applied to a trained tensor-compressed model for further model size reductions.

\vspace{-10pt}
\section{Performance Optimization of CoMERA}
\vspace{-5pt}
\label{sc:optimization}
While CoMERA can greatly reduce training variables and memory cost, the low-rank and small-size tensor operations in CoMERA are not efficiently supported by GPU. This often slows the training process. This section presents three methods to achieve real training speedup on GPU.

\subsection{Performance Optimization of TTM Embedding Tables.}
\label{sc:embedding}
\begin{wrapfigure}{r}{2.5in}
         \centering
         \includegraphics[width=2.5in]{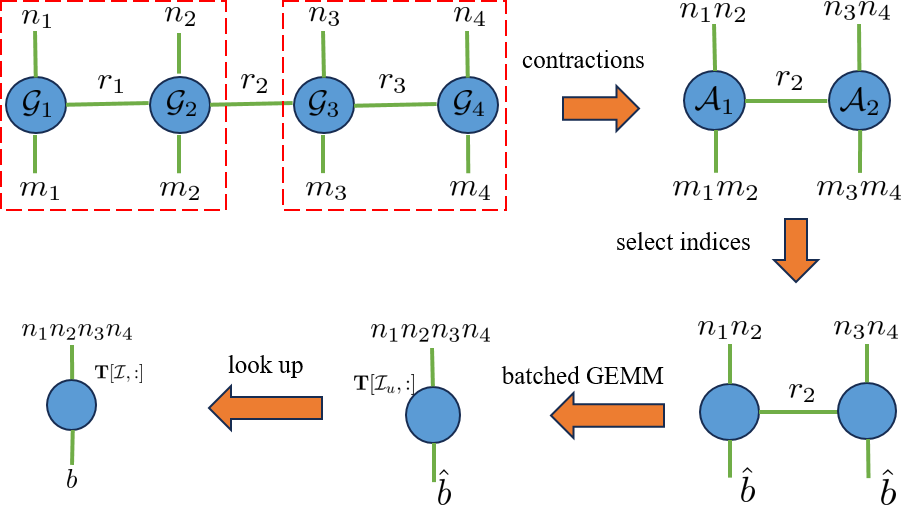}
    \caption{Optimized TTM embedding lookup.}
    \vspace{-15pt}
         \label{fig:TTM-emb}
\end{wrapfigure}
Embedding tables are widely used to transfer discrete features into continuous hidden space. The row size of embedding tables is usually much larger than the column size, making TTM compression more suitable than the TT format. In the following, we use an order-4 TTM embedding table to illustrate how to accelerate the lookup process. 

%\zz{Need to improve the readability of this subsection. It's still hard to understand the optimized look-up process.}

We consider an embedding table $\mat{T}\in \mathbb{R}^{m\times n}$. A look-up operation selects the submatrix $\mat{T}[\cI,:]\in \mathbb{R}^{b\times n}$ for the index set $\cI=\{i_k\}_{k=1}^b$. This operation is fast and inexpensive. However, the full embedding table itself is extremely memory-consuming. Suppose that $m=m_1m_2m_3m_4, n=n_1n_2n_3n_4$, then we reshape $\mat{T}$ into tensor $\ten{T}\in \mathbb{R}^{m_1\times n_1\times \cdots \times m_4 \times n_4}$ and represent it in TTM format
\begin{equation}
    \ten{T} = \ten{G}_1 \times_{4,1} \ten{G}_2 \times_{4,1} \ten{G}_3 \times_{4,1} \ten{G}_4.
\end{equation}
The compressed embedding table does not have the matrix $\mat{T}$ explicitly. We convert each row index $i_k \in \cI$ to a tensor index vector $(z_1^k,z_2^k,z_3^k,z_4^k)$ and denote $\mathcal{Z}_t = \{z_t^k\}_{k=1}^b$, then $\mat{T}[\cI,:]$ can be computed by contracting the tensors $\{\ten{G}_t[:,\mathcal{Z}_t,:,:]\}_{t=1}^4$ where each has size $r_{t-1}\times b \times n_i \times n_t$. The $\ten{G}_t[:,\mathcal{Z}_t,:,:]$ stores many duplicated values especially when the set $\cI$ is large. Therefore, directly computing the tensor contractions can cause much computing and memory overhead. 

We optimize the tensor contraction by eliminating the redundant computation at two levels. {\bf Row-index level.} We construct the index set $\cI_{u}=\{i_k\}_{k=1}^{\hat {b}}$ containing all unique indices in $\cI$.  We can easily obtain $\mat{T}[\cI,:]$ from $\mat{T}[\cI_u,:]$. {\bf Tensor-index level.} The reduced index set $\cI_{u}$ leads to $\hat{b}$ associated tensor index vectors $(z_1^k,z_2^k,z_3^k,z_4^k)$, but at most $m_1m_2$ pairs of $(z_1^k,z_2^k)$ and $m_3m_4$ pairs of $(z_3^k,z_4^k)$ are unique.  For instance, (2,3,1,3) and (2,3,2,4) are common in (2,3), so we only compute (2,3) entry once. Therefore, we can consider all unique pairs $(z_1^k,z_2^k)$ and $(z_3^k,z_4^k)$ and compute 
    \begin{align}
    \label{eq:emb_A1}
    \ten{A}_1 = &\einsum(r_0m_1n_1r_1,r_1m_2n_2r_2\Rightarrow (m_1m_2)(n_1n_2)r_2,[\ten{G}_1,\ten{G}_2]),\\
    \label{eq:emb_A2}
    \ten{A}_2 = & \einsum(r_2m_3n_3r_3,r_3m_4n_4r_4\Rightarrow r_2(m_3m_4)(n_3n_4),[\ten{G}_3,\ten{G}_4]).
    \end{align}
For each $i_k\in \cI_u$, let $(j_1^k,j_2^k)$ be the coordinate of $i_k$ for size $(m_1m_2,m_3m_4)$. We denote $\mathcal{J}_1=\{j_1^k\}_{k=1}^{\hat{b}}$ and $\mathcal{J}_2=\{j_2^k\}_{k=1}^{\hat{b}}$, then compute the unique rows of $\mat{T}$ as  
\begin{equation}
    \label{eq:emb_unique}
    \mat{T}[\cI_u,:] = \einsum(\hat{b}(n_1n_2)r_2,r_2 \hat{b}(n_3n_4)\Rightarrow \hat{b}(n_1n_2n_3n_4),[\ten{A}_1[\mathcal{J}_1,:,:],\ten{A}_1[:,\mathcal{J}_2,:]]).
\end{equation}
Figure \ref{fig:TTM-emb} summarizes the whole process of TTM embedding table look-up. This approach can be easily applied to higher-order embedding tables by first grouping some small tensor cores to obtain intermediate tensors and then utilizing them to compute unique row vectors.

\begin{comment}
\begin{figure}[t]
    \centering
    \includegraphics[width=0.7\textwidth]{figures/TTM_emb_new.png}
    \caption{TTM embedding table lookup}
    \label{fig:TTM-emb}
\end{figure}
\end{comment}

\begin{figure}[t]
     \centering
     \begin{subfigure}[b]{0.48\textwidth}
         \centering
         \includegraphics[width=\textwidth]{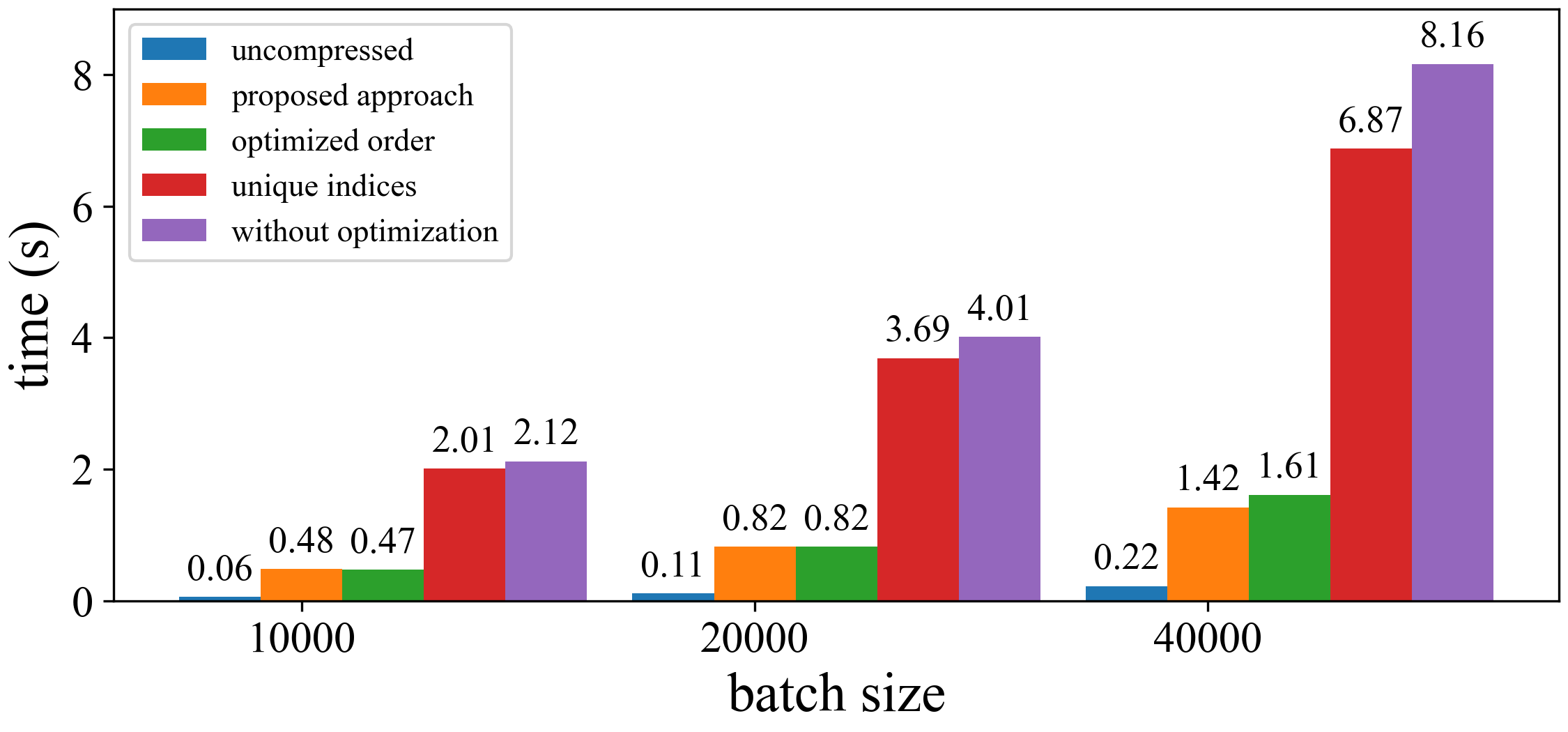}
         \caption{Speed-up}
     \end{subfigure}
     \hfill
     \begin{subfigure}[b]{0.48\textwidth}
         \centering
         \includegraphics[width=\textwidth]{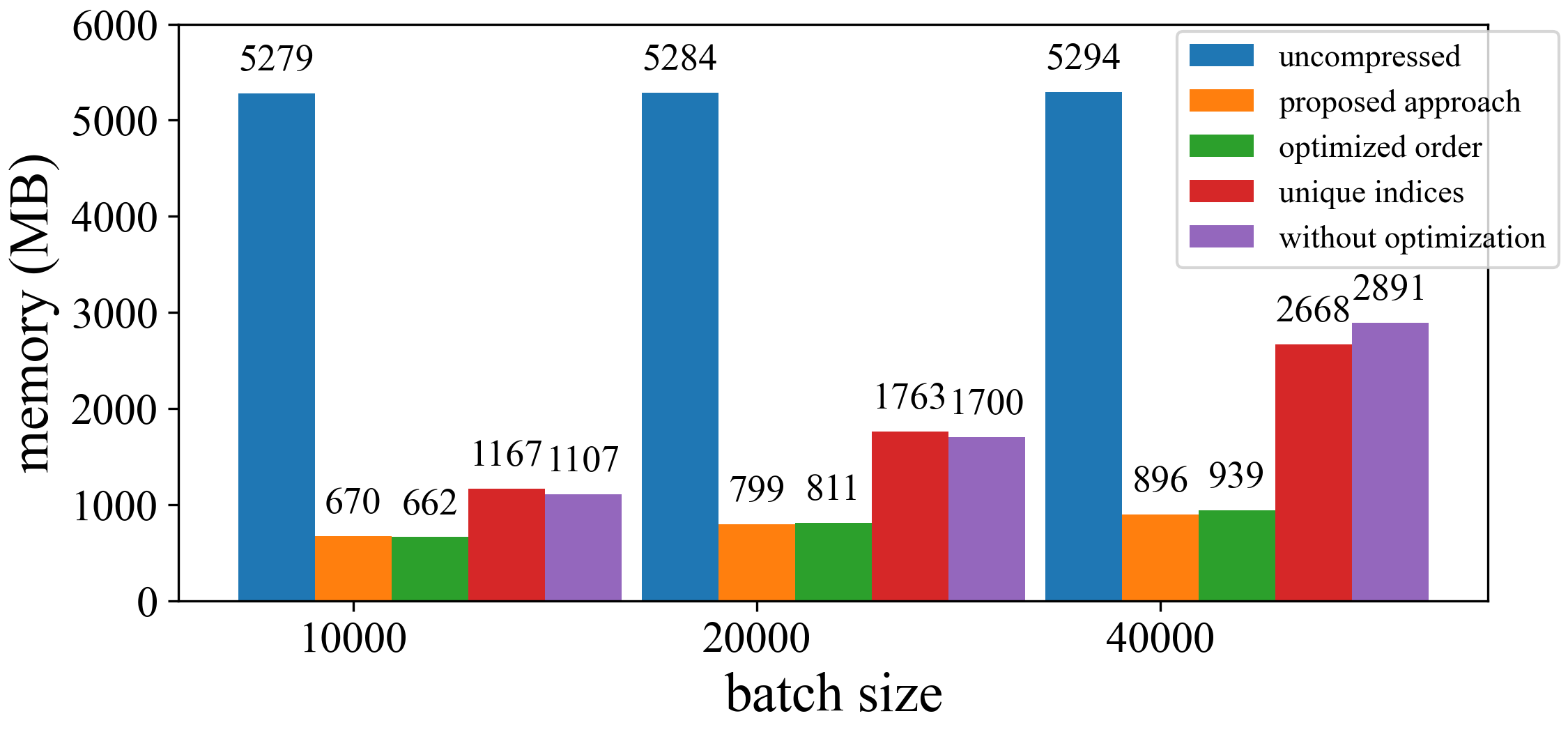}
         \caption{Memory reduction}
     \end{subfigure}
     \caption{Performance of optimized TTM embedding table lookup. The labels \textit{uncompressed, proposed approach, optimized order, unique indices, without optimization} represent standard embedding with sparse gradients, the new method in \ref{sc:embedding}, the method that only uses the unique order, the method that only uses the unique indices, and the method without optimization, respectively.}
     \label{fig:emb_TTM_speed}
     \vspace{-15pt}
\end{figure}

 % For the TTM embedding table with tensor cores $\ten{G}_1,\ldots,\ten{G}_6$, we may first compute $\ten{G}_1\times_{4,1} \ten{G}_2,\ten{G}_3\times_{4,1} \ten{G}_4, \ten{G}_5\times_{4,1} \ten{G}_6$ or $\ten{G}_1\times_{4,1} \ten{G}_2 \times_{4,1} \ten{G}_3,\ten{G}_4\times_{4,1} \ten{G}_5 \times_{4,1} \ten{G}_6$. The choice depends on the shapes of the tensor cores. 
\vspace{-10pt}
\paragraph{Performance.} We demonstrate the optimized TTM embedding tables on a single RTX 3090 GPU. We consider an embedding table of TTM shape $[[80,50,54,50],[4,4,4,2]]$ and rank $32$, extracted from a practical DLRM model. As shown in Figure \ref{fig:emb_TTM_speed}, our proposed method achieves about $4-5\times$ speed-up and $2-3\times$ memory saving than the standard TTM embedding without any optimization. The uncompressed embedding with sparse gradients is faster than our approach since our TTM embedding table requires extra computation, but it uses much more memory than the TTM embedding table.

%The acceleration and memory reduction mainly come from the optimized contraction order. Using the unique indices brings further performance improvement. % in addition to the optimized computation.

% \begin{table}[]
%     \centering
%     \caption{Speed-up of optimized embedding}
%     \label{tab:emb}
%     \begin{tabular}{|c|c|c|c|c|c|c|c|c|}
%     \hline
%        \multirow{2}{*}{batch} &  \multicolumn{4}{c|}{time (s)} &  \multicolumn{4}{c|}{memory (MB)} \\
%        \cline{2-9}
%           & new & unique & order & old & new & unique & order & old \\
%         \hline
%        10000 & 0.9 & 3.3 & 1.1 & 4.3 & 775 & 1360& 785 &  1394 \\
%        20000 & 1.6 & 6.4& 1.9 &8.8 & 939 &  2048  &1022 & 2190 \\
%        40000 & 3.1 & 12.2 & 3.6 & 17.8 & 1220 &3265 &1342 &3784 \\
%        \hline
%     \end{tabular}
% \end{table}

%\zy{will add one more paragraph about memory/flop reduction}

\subsection{Contraction Path Optimization for TT-Vector Multiplications}
\label{subsec:path}
Next, we optimize the forward- and back- propagation of linear layers in the TT format. We consider the linear layer $\mat{Y}=\mat{X}\mat{W}$, where $\mat{Y}\in\mathbb{R}^{b\times N_2}, \mat{W}\in \mathbb{R}^{N_1\times N_2},\mat{X}\in \mathbb{R}^{b\times N_1}$. The $\mat{W}$ is compressed into the Tensor-Train format: $\ten{W}=[\![\ten{G}_1,\ldots,\ten{G}_{2d}]\!] \in \mathbb{R}^{n_1\times \cdots \times n_{2d}}$,where $\ten{G}_i\in \mathbb{R}^{r_{i-1}\times n_i \times r_{i}}$ and $N_1=n_1\cdots n_d,N_2=n_{d+1}\cdots n_{2d}$. The forward-propagation in the einsum form is
\begin{equation} \label{eq:forward}
    \mat{Y}=\mat{X}\mat{W} = \einsum (bn_1\ldots n_d,S_1,\ldots,S_{2d}\Rightarrow bn_{d+1}\ldots n_{2d},[\ten{X},\ten{G}_1,\ldots,\ten{G}_{2d}])
\end{equation}
where $\ten{X}\in \mathbb{R}^{b\times n_1\times \cdots \times n_d}$ is the reshaping of $\mat{X}$ and $S_i$ denotes $r_{i-1}n_ir_i$. Suppose that the gradient to $\mat{Y}$ is $\mat{g}_\mat{Y}$, then the gradients to $\ten{G}_i$ and $\ten{X}$ can be computed as follows:
\begin{align} \label{eq:back Gi}
    \mat{g}_{\ten{G}_i}& = \einsum \big(bn_1\ldots n_d,bn_{d+1}\ldots n_{2d},S_1,\ldots,S_{i-1},S_{i+1},\ldots,S_{2d}\Rightarrow S_{i}, \nonumber\\ 
    & \quad \quad \quad \quad \quad \quad \quad \quad \quad \quad \quad \quad \quad\,[\ten{X},\mat{g}_\ten{Y},\ten{G}_1,\ldots,\ten{G}_{i-1},\ten{G}_{i+1},\ldots,\ten{G}_{2d}]),\\ 
 \label{eq:back X}
    \mat{g}_{\ten{X}} & = \einsum \big(bn_{d+1}\ldots n_{2d},S_1,\ldots,S_{2d}\Rightarrow bn_1\ldots n_d,
    [\mat{g}_\ten{Y},\ten{G}_1,\ldots,\ten{G}_{2d}]).
\end{align}
In total, $2d+2$ contraction sequences are needed for the TT-format forward- and back- propagation. To reduce the computational costs, it is critical to find an optimal or near-optimal contraction path. 

\vspace{-10pt}
\paragraph{Large batch case.} We denote $\ten{A}_i:=\ten{G}_1 \times \cdots \times \ten{G}_i, \ten{A}_{-i}=\ten{G}_{d-i+1}\times \cdots \times \ten{G}_d,
    \ten{B}_i:=\ten{G}_{d+1} \times \cdots \times \ten{G}_{d+i}, \ten{B}_{-i}=\ten{G}_{2d-i+1}\times \cdots \times \ten{G}_{2d}$,
% \begin{align}
%     \label{eq:path-Ai}
%     &\ten{A}_i:=\ten{G}_1 \times \cdots \times \ten{G}_i, &\ten{A}_{-i}=\ten{G}_{d-i+1}\times \cdots \times \ten{G}_d, \\
%     &\ten{B}_i:=\ten{G}_{d+1} \times \cdots \times \ten{G}_{d+i}, & \ten{B}_{-i}=\ten{G}_{2d-i+1}\times \cdots \times \ten{G}_{2d}
% \end{align}
which are all computed sequentially. In practice, we only need to compute $\ten{A}_d,\ten{A}_{-d},\ten{B}_d,\ten{B}_{-d}$ and store the intermediate results. The forward-propagation \eqref{eq:forward} is then computed in the following way 
\begin{eqnarray}
    \label{eq:path-T1}
    \ten{T}_1 &=& \einsum(bn_1\ldots n_d,n_1\ldots n_d r_d \Rightarrow br_d,[\ten{X},\ten{A}_d]) \\
    \label{eq:path-Y}
    \ten{Y} &=& \einsum(br_d,r_dn_{d+1}\ldots n_{2d} \Rightarrow bn_1\ldots n_d,[\ten{T}_1,\ten{B}_d]).
\end{eqnarray}
In backward propagation, the gradients are computed in the following way:
\vspace{-5pt}
\begin{itemize}[leftmargin=20pt]
    \item The gradient $\mat{g}_\ten{X}$ is computed as 
\begin{eqnarray}
    \label{eq:path-U1}
    \ten{U}_1 &=& \einsum (bn_{d+1}\ldots n_{2d},r_dn_{d+1}\ldots n_{2d} \Rightarrow br_d,[\mat{g}_{\ten{Y}},\ten{B}_d])\\
    \label{eq:path-gX}
    \mat{g}_\ten{X} &=& \einsum(br_d,n_{1}\ldots n_{d}r_d \Rightarrow bn_1\ldots n_d,[\ten{U}_1,\ten{A}_d]).
\end{eqnarray}

\vspace{-4pt}
\item The gradients $\mat{g}_{\ten{G}_i}$ for $i\ge d+1$ can be computed as 
\begin{eqnarray}
\label{eq:path-T2}
     \ten{T}_2 &=& \einsum (br_d,bn_{d+1}\ldots n_{2d} \Rightarrow r_dn_{d+1}\ldots n_{2d},[\ten{T}_1,\mat{g}_{\ten{Y}}])\\
     \label{eq:path-gGi>d}
     \mat{g}_{\ten{G}_i} &=& \einsum (r_dn_{d+1}\ldots n_{2d},r_dn_{d+1}\ldots n_{i-1} r_{i-1}, r_{i} n_{i+1} \ldots n_{2d} \Rightarrow\\
     &&  r_{i-1}n_ir_{i}, \;[\ten{T}_2,\ten{B}_{i-1-d},\ten{B}_{-(2d-i)}] ). \nonumber
\end{eqnarray}

\vspace{-4pt}
\item Similarly, the gradients $\mat{g}_{\ten{G}_i}$ for $i\le d$ can be computed as
\begin{eqnarray}
\label{eq:path-U2}
     \ten{U}_2 &=& \einsum (br_d,bn_1\ldots n_d \Rightarrow r_d n_1 \ldots n_d,[\ten{U}_1,\ten{X}]) \\
     \label{eq:path-gGi<d}
     \mat{g}_{\ten{G}_i} &=& \einsum (r_dn_{1}\ldots n_{d},n_{1}\ldots n_{i-1} r_{i-1}, r_{i} n_{i+1} \ldots n_{d} r_d \Rightarrow \\
     &&  r_{i-1}n_ir_{i}, \;[\ten{U}_2,\ten{A}_{i},\ten{A}_{-(d-i)}] ). \nonumber
\end{eqnarray}
\end{itemize}
The contraction paths of forward- and back- propagation are summarized in Appendix \ref{sec:contract-algorithm}.

\vspace{-10pt}
\paragraph{Analysis.} 
The proposed empirical path is near-optimal for large batch sizes. The following result analyzes the contraction path for forward-propagation. 

\begin{proposition} \label{prop:forward}
Suppose that the TT ranks satisfy $1=r_0< r_1 \le \cdots \le r_d \ge r_{d-1} > \cdots \ge r_{2d}=1$ and the batch size $b$ is large enough. There exist groups $\{S_i\}_{i=1}^k$ where $S_i=\{\ten{G}_{j_i+1},\ldots,\ten{G}_{j_{i+1}}\}$ containing consecutive tensor cores for $0= j_1<\cdots <j_k<j_{k+1}=2d$. Then, the contraction path with the least number of flops for the forward-propagation \eqref{eq:forward} first contracts the tensor cores in each $S_i$ to obtain $\ten{V}_i$ with dimension $r_{j_i}\times n_{j_i+1} \times \cdots \times n_{j_{i+1}}\times r_{j_{i+1}}$ and then contract the input tensor $\ten{X}$ with tensors $\{\ten{V}_i\}_{i=1}^k$ in the sequential order.
\end{proposition}

% \begin{proposition}[Informal] \label{prop:forward}
%     Suppose that the batch size $b$ is large enough, then the contraction path with the least number of flops for the forward-propagation \eqref{eq:forward} first contracts some groups of adjacent tensor cores and then contracts the input tensor with them in the sequential order. (See Appendix \ref{appendix:path proof} for the formal statement and the complete proof.) \zz{The proposition should be formal, but it should also be easy to understand. I know this is challenging...}
% \end{proposition}

\vspace{-10pt}
\begin{proof}
    See Appendix \ref{appendix:path proof} for the complete proof.
\end{proof}
\vspace{-10pt}

Proposition \ref{prop:forward} implies that the optimal path first contracts some consecutive tensor cores and then contracts obtained tensors with the input tensor sequentially. The groups $\{S_i\}_{i=1}^k$ depend on the dimensions, ranks, and batch size. The proposed contraction path satisfies the property shown in Proposition \ref{prop:forward} and has flops roughly $b(n_1\cdots n_d+n_{d+1}\cdots n_{2d})r_d$. The optimal contraction path has flops about $bn_1\cdots n_dc_1+bn_{d+1}\cdots n_{2d}c_2$, where $c_1,c_2$ are some constants. Hence, the proposed is near-optimal and has a comparable complexity to the optimal path. Suppose the optimal path is different from the proposed empirical path. Then the optimal path will likely involve a few more large intermediate tensors, which pose more memory costs during training and inference especially for static computational graphs. The empirical path is a good choice to balance time and memory consumption. Similar arguments can be applied to the contractions for back-propagation. 

When the batch size is small, the optimal path may have much fewer flops. However, the execution time is almost the same as the proposed path since all the operations are too small. Hence, we can use the proposed path for most batch sizes. See Appendix \ref{appendix:path} for more analysis.

\subsection{GPU Performance Optimization via CUDA Graph}
%\subsubsection{CUDA Graphs for Reducing Tensor Computation Overhead}

While CoMERA consumes much less computing FLOPS than standard uncompressed training, it can be slower on GPU if not implemented carefully. Therefore, it is crucial to optimize the GPU performance to achieve real speedup. Modern GPUs are highly optimized for large-size matrix multiplications. However, the small-size tensor contractions in CoMERA are not yet optimized on GPU and require many small-size GPU kernels, causing significant runtime overhead. During the training, Cuda Graph launches and executes the whole computing graph rather than launching a large number of kernels sequentially. This can eliminate lots of back-end overhead and lead to significant training speedup. It is more suitable for CoMERA since tensor-compressed training has much more small kernels than uncompressed training. This is just an initial step of GPU optimization. We expect that a more dedicated GPU optimization can achieve a more significant training speedup.

\begin{table}[t]
    \caption{Result of Transformer on MNLI of batch size 128.}
    \label{tab:MNLI-compression}
    \centering \small
    \begin{tabular}{|c|c|c|c|}
    \hline
          & validation & total size (MB) & compressed size (MB) \\
         \hline
        uncompressed training & 62.2\% & 256 ($1\times$) &253 ($1\times$)\\ 
        CoMERA (early stage)  & 63.3\% & 5.9 ($43\times$) & 3.4 ($74\times$)\\
        CoMERA (late stage), target ratio: 0.8 &  62.2\% & 4.9 ($52\times$) & 2.4 ($105\times$)\\
         CoMERA (late stage), target ratio: 0.5 &  62.1\% & 3.9 ($65\times$) & 1.4 ($181\times$)\\
         CoMERA (late stage), target ratio: 0.2 &  61.5\% & 3.2 ($80\times$) & 0.7 ($361\times$)\\
        \hline
    \end{tabular} \normalsize
    \vspace{-15pt}
\end{table}

\vspace{-10pt}
\section{Training Results}\label{sc:results}
\vspace{-5pt}
In this section, we test the performance of CoMERA on a few benchmarks, including a domain-specific LLM. Our experiments are run on a Nvidia RTX 3090 GPU with 24GB RAM. 

\vspace{-10pt}

\subsection{A Medium-Size Transformer with Six Encoders}
\label{sc:6encoder}

\begin{wrapfigure}{r}{2.7in}
    \centering
     \vspace{-10pt}
     \includegraphics[width=0.45\textwidth]{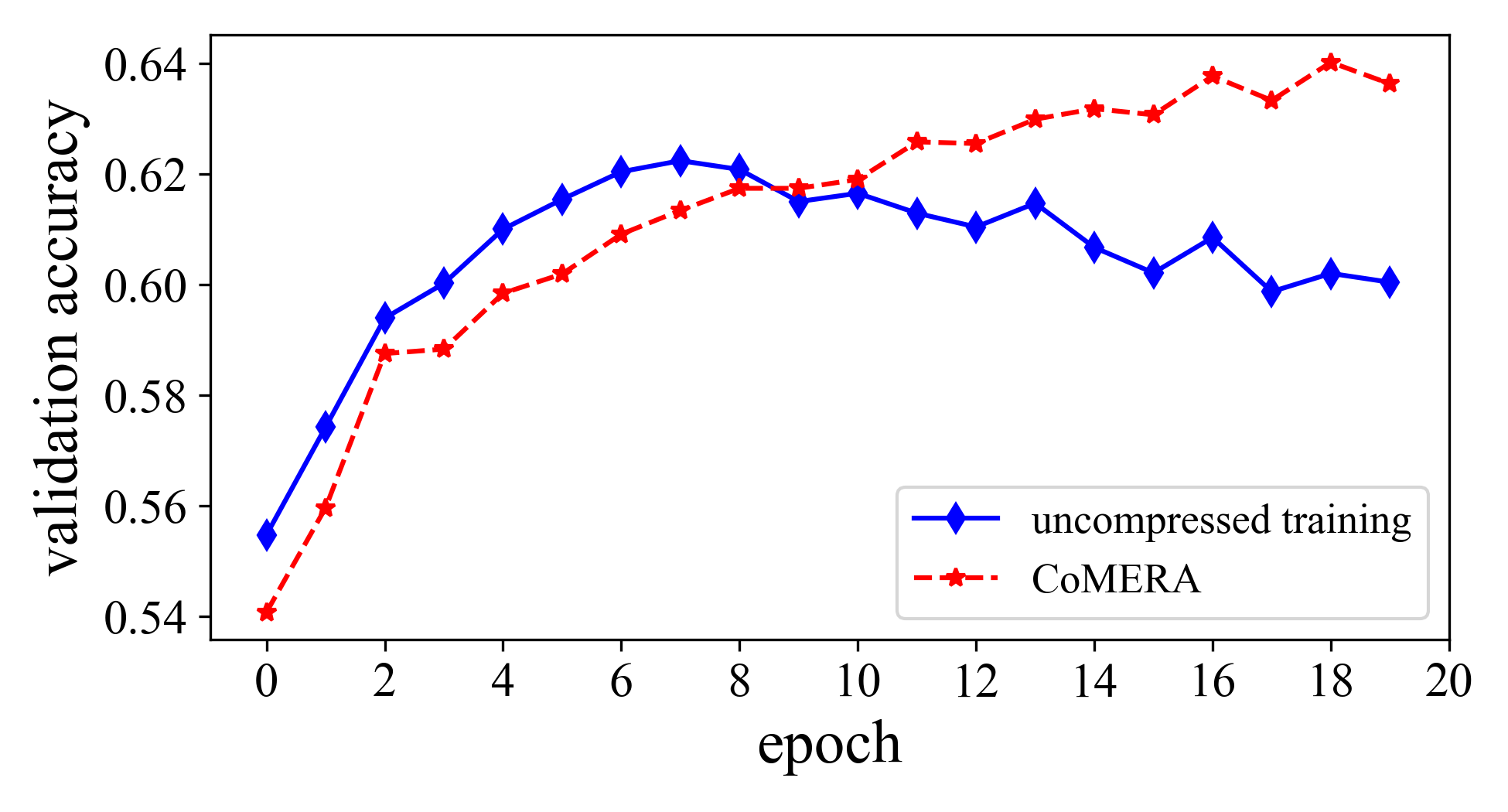}
     \caption{Behavior of early-stage CoMERA training on the MNLI dataset.}
     \label{fig:MNLI_val}
     \vspace{-15pt}
\end{wrapfigure}

%\vspace{-5pt}
We first consider a six-encoder transformer. The embedding tables and all linear layers are represented as tensor cores in the training process as detailed in Appendix \ref{appendix:settings}. We train this model on the MNLI dataset \cite{N18-1101} with the maximum sequence length $128$ and compare the accuracy, resulting model size, and training time of CoMERA with the standard uncompressed training.

\begin{table}[t]
    \centering
     \caption{The change of ranks of layers in the fifth encoder block.}
    \label{tab:ranks}
    \small
    \begin{tabular}{|c|c|c|c|}
    \hline
      &  before training & early-stage rank & late-stage rank \\
        \hline
         Q-layer in attention & $(12,30,30,30,12)$ & $(12,30,30,30,12)$ & $(0,0,0,0,0)$\\
         \hline
         K-layer in attention & $(12,30,30,30,12)$ & $(12,30,30,30,12)$ & $(0,0,0,0,0)$\\
         \hline
         V-layer in attention & $(12,30,30,30,12)$ & $(12,30,30,30,12)$ & $(9, 11, 11, 7, 9)$\\
         \hline
         FC-layer in attention & $(12,30,30,30,12)$ & $(12,30,29,30,12)$ & $(9, 8, 10, 8, 8)$\\
         \hline
         \#1 linear-layer in Feed-Forward &$(12,30,30,30,16)$ & $(0,0,0,0,0)$ & $(0,0,0,0,0)$\\
         \hline
         \#2 linear-layer in Feed-Forward & $(16,30,30,30,12)$ & $(0,0,0,0,0)$ & $(0,0,0,0,0)$\\
         \hline
    \end{tabular} \normalsize
    \vspace{-10pt}
\end{table}

\vspace{-10pt}
\paragraph{CoMERA Accuracy and Compression Performance.} Table \ref{tab:MNLI-compression} summarizes the training results. The {\bf early-stage training} of CoMERA achieves $74\times$ compression ratio on all tensorized layers, and the validation accuracy is even higher than the uncompressed training. Figure \ref{fig:MNLI_val} shows the validation accuracy of CoMERA. In the {\bf late stage} of CoMERA, we set different target compression ratios for more aggressive rank pruning. The target compression ratios are for the tensorized layers rather than for the whole model. The late-stage training can reach the desired compression ratio with very little accuracy drop. The smallest model has a compression ratio of $80\times$ for the whole model due to a $361\times$ compression on the tensorized layers with slightly worse accuracy. 

\begin{wrapfigure}{r}{2.7in}
         \centering
         \vspace{-5pt}
         \includegraphics[width=2.7in]{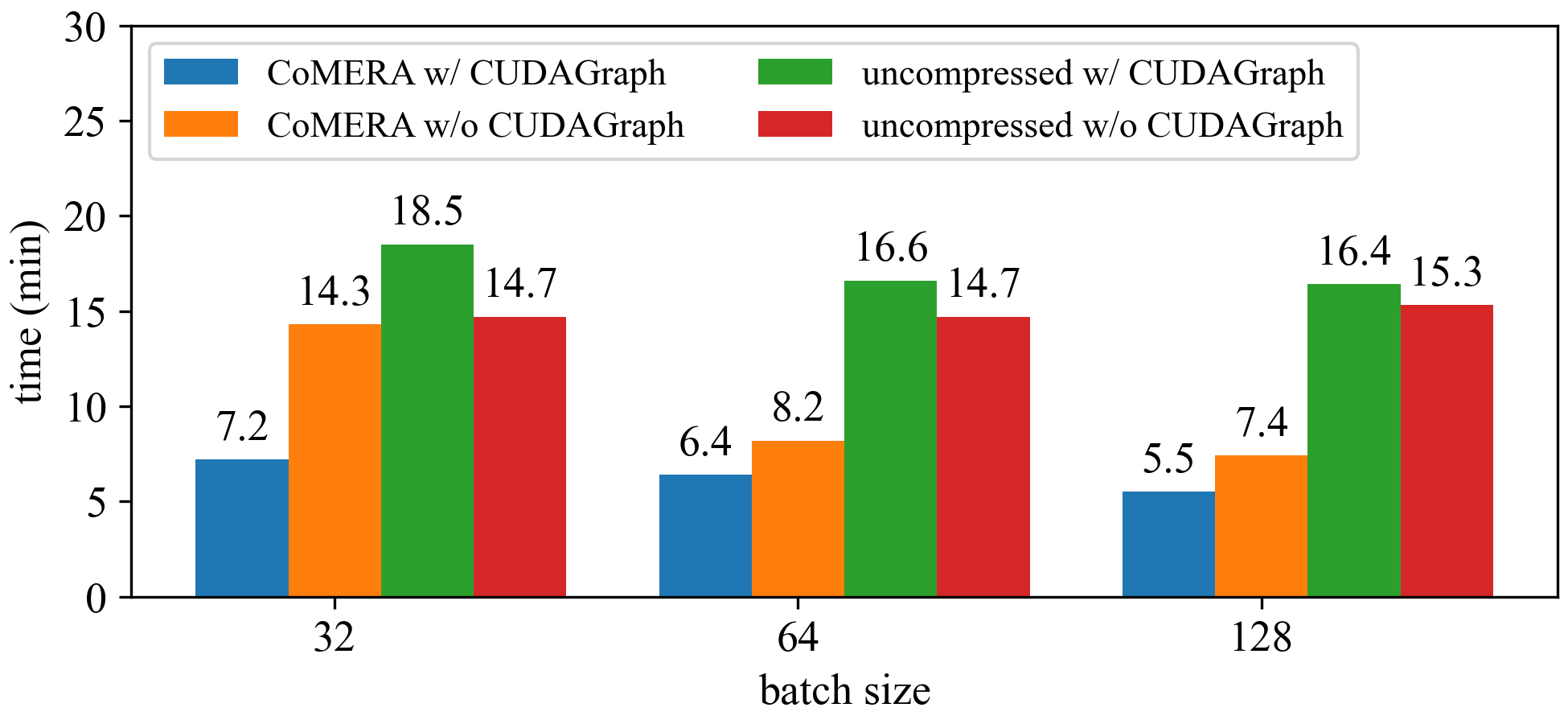}
     \caption{Training time per epoch for the six-encoder transformer model on the MNLI dataset.}
         \label{fig:MNLI-time}
         \vspace{-15pt}
\end{wrapfigure}

\vspace{-10pt}
\paragraph{Architecture Search Capability of CoMERA.} A major challenge in training is architecture search: shall we keep certain layers of a model? Interestingly, CoMERA has some capability of automatic architecture search. Specifically, the ranks of some layers become zero in the training, and thus the whole layer can be removed. For the target compression ratio $0.2$, the whole second last encoder and some linear layers in other encoders are completely removed after late-stage rank-adaptive training. The change of ranks of layers in the 5th encoder is shown in Table \ref{tab:ranks}.

\begin{comment}
\begin{figure}[t]
    \centering
    \includegraphics[width=0.8\textwidth]{figures/CoMERA_MNLI_time.png}
    \caption{Training time per epoch for the six-encoder transformer model on the MNLI dataset.}
    \label{fig:MNLI-time}
\end{figure}
\end{comment}

% \begin{figure}
%     \centering
%     \includegraphics[width=0.6\textwidth]{MNLI_val_curve.png}
%     \caption{Validation accuracy curve for tensor-compressed training on MNLI dataset}
%     \label{fig:MNLI_val}
% \end{figure}

% \begin{figure}
%     \centering
%     \includegraphics[width=0.6\textwidth]{MNLI_size_acc.png}
%     \caption{Validation accuracy curve for tensor-compressed training on MNLI dataset}
%     \label{fig:MNLI_size_acc}
% \end{figure}

\vspace{-10pt}
\paragraph{Training Time.} As shown in Figure \ref{fig:MNLI-time}, CoMERA with CUDAGraph achieves around $3\times$ speed-up than uncompressed training. CoMERA without CUDAGraph can take much longer time in small batch-size setting due to the launching overhead of too many small kernels. The uncompressed training with CUDAGraph takes longer time than the one without CUDAGraph. This is because CUDAGraph requires all batches to have the same sequence length, and the consequent computing overhead is more than the time reduction of CUDAGraph. In contrary, CoMERA has much fewer computing FLOPS and the computing part accounts for a much smaller portion of the overall runtime. Empirically CoMERA is $2-3\times$ faster in the whole training than uncompressed training for transformers on a single GPU, but we do not have theoretical guarantees about the number of epochs although they are similar in our experiments. Appendix \ref{appendix:training_MNLI_ratios} provides more details about the run-time comparison on this benchmark, showing that CoMERA is still faster than standard training even if the compression ratio is close to 1.

% \begin{table}[t]
%     \centering
%     \caption{Training time per epoch for the six-encoder transformer model on the MNLI dataset.}
%     \label{tab:MNLI-time}
%     \begin{tabular}{|c|c|c|c|c|}
%     \hline
%          \multicolumn{2}{|c|}{batch size}  & 32 &64 & 128 \\
%          \hline
%         \multirow{2}{*}{CUDAGraph} & CoMERA  & 7.2 min & 6.4 min & 5.5 min\\
%                                 & uncompressed & 18.5 min & 16.6 min & 16.4 min\\
%                                 \hline 
%         \multirow{2}{*}{w/o CUDAGraph} & CoMERA & 14.3 min & 8.2 min & 7.4 min\\
%                                 & uncompressed & 14.7 min & 14.7 min & 15.3 min\\
%         \hline
%     \end{tabular}
    
% \end{table}

\vspace{-10pt}
\subsection{A DLRM Model with 4-GB Model Size}
\vspace{-5pt}

We further test CoMERA on DLRM \cite{naumov2019deep} released by Meta on Criteo Ad Kaggle dataset \cite{criteo-display-ad-challenge}. We compress the ten largest embedding tables into the TTM format as in Section \ref{sc:embedding}. All fully connected layers with sizes $>128$ are compressed into TT format. The model is trained for two epochs. 
\begin{wrapfigure}{r}{2.3in}
         \centering
         \vspace{-0pt}
         \includegraphics[width=2.3in]{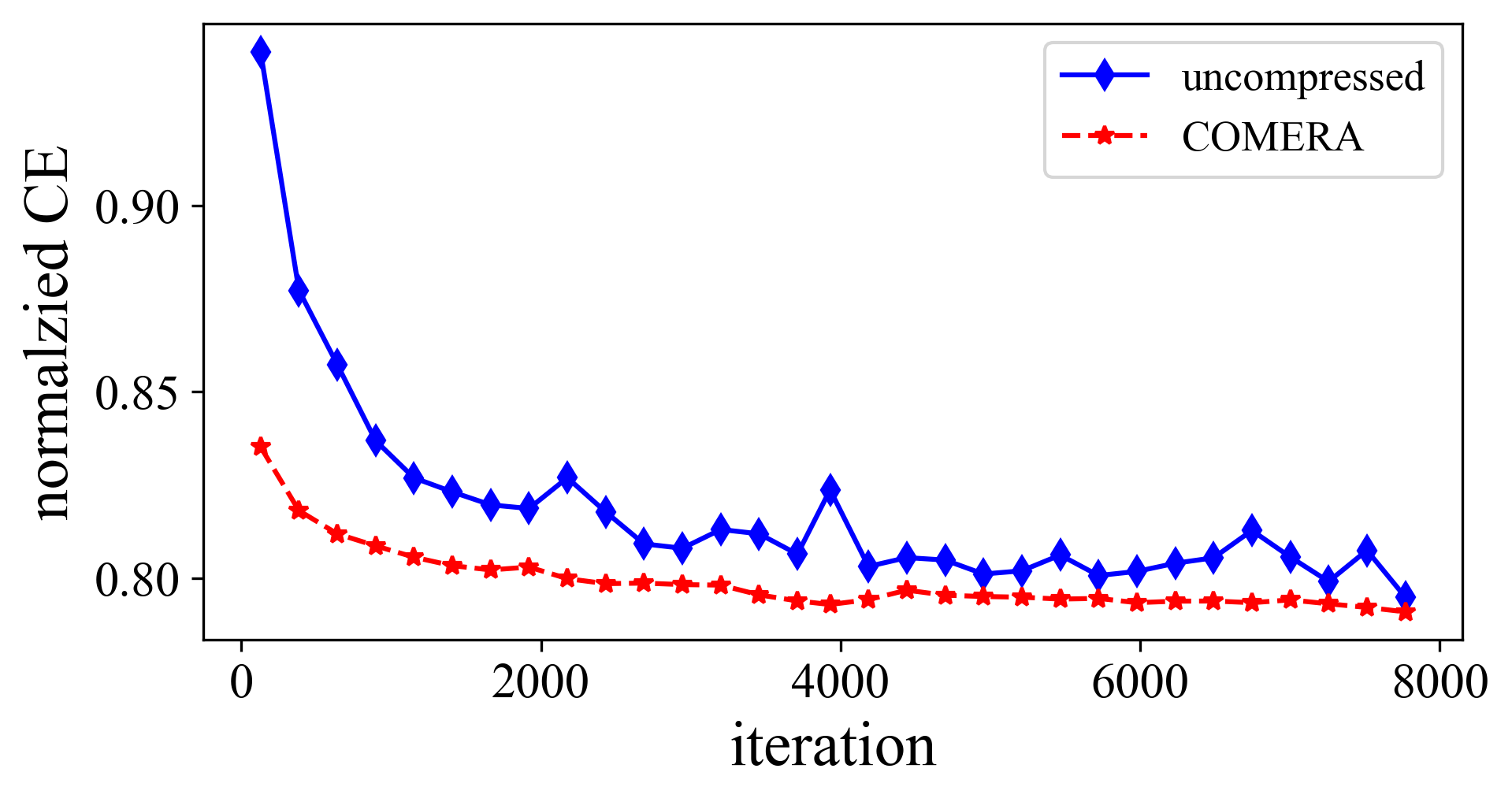}
     \caption{NCE loss curve of DLRM on the validation dataset.}
         \label{fig:DLRM_convergence}
         \vspace{-15pt}
\end{wrapfigure}

\begin{figure}
     \centering
     \vspace{-15pt}
     \begin{subfigure}[b]{0.44\textwidth}
         \centering
         \includegraphics[width=\textwidth]{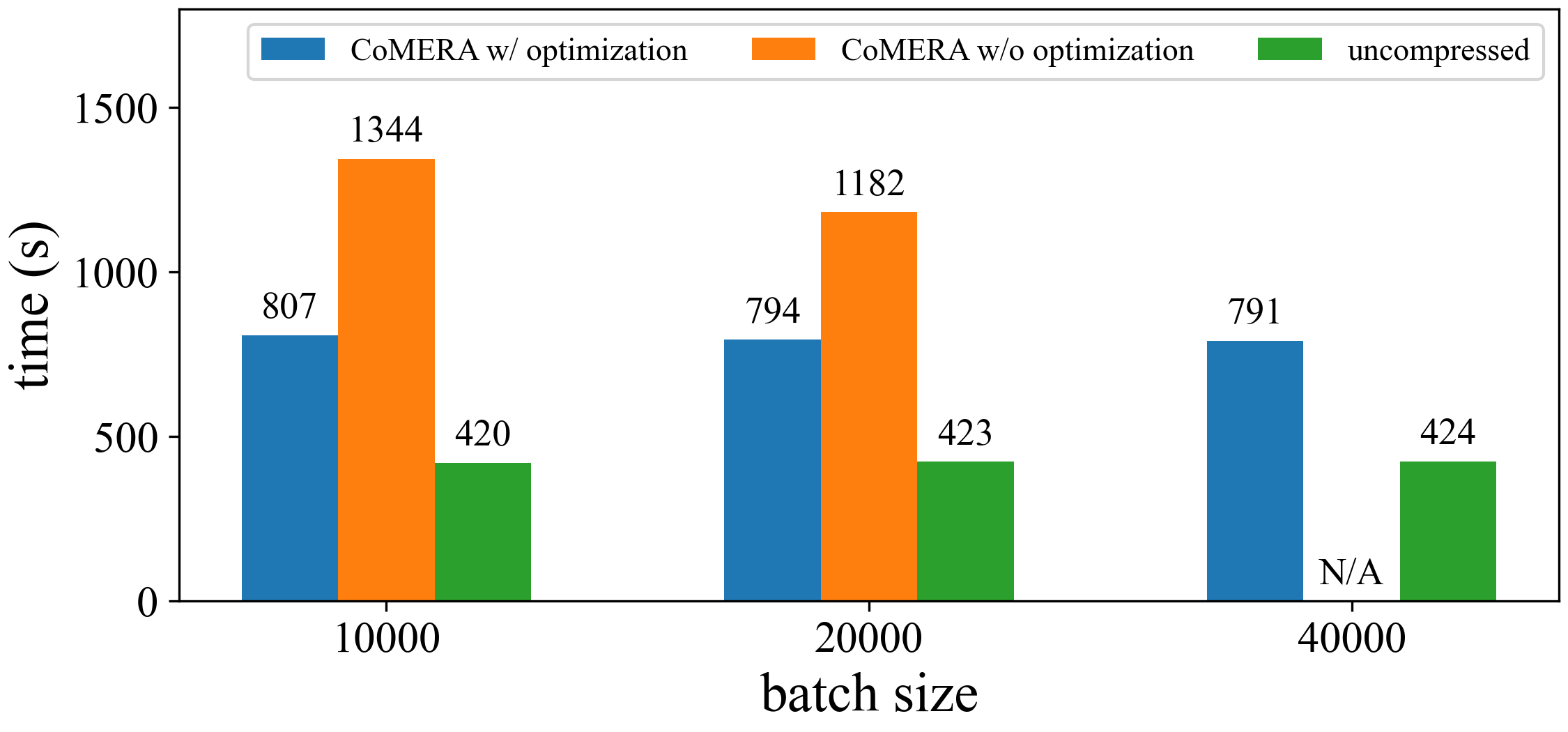}
         \caption{Training time per epoch.}
     \end{subfigure}
     \hfill
     \begin{subfigure}[b]{0.44\textwidth}
         \centering
         \includegraphics[width=\textwidth]{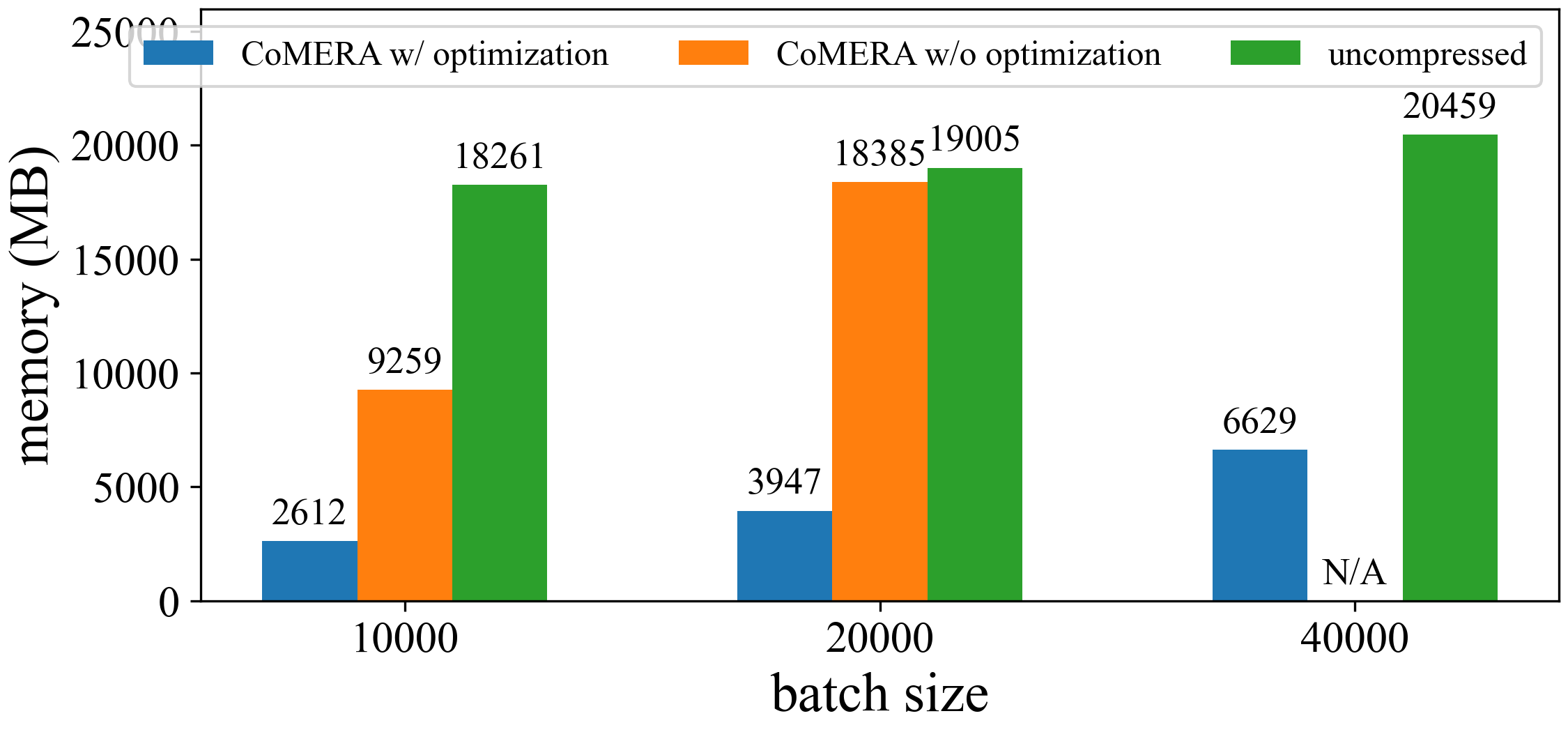}
         \caption{Memory cost.}
     \end{subfigure}
     \caption{Performance of optimized CoMERA on training DLRM.}
     \vspace{-15pt}
     \label{fig:emb-DLRM}
\end{figure}

\vspace{-10pt}
\paragraph{Effect of Optimized TTM Embedding.} The training time per epoch and peak memory cost are shown in Figure \ref{fig:emb-DLRM}. Our optimized TTM lookup speeds up the training process by around $2\times$ and remarkably reduces the memory cost by $4-5\times$. %, especially for large batch training. 

\vspace{-10pt}
\begin{wraptable}{r}{3in}
    \centering
    \vspace{-10pt}
    \caption{Training results on the DLRM model with a batch size $10,000$. }
    \label{tab:train-DLRM} \small
    \begin{tabular}{|c|c|c|}
    \hline
    & uncompressed & CoMERA \\
    \hline
   accuracy & 78.68\% &  78.76\% \\
    normalized CE& 0.793 & 0.792 \\
   model size (GB) & 4.081 & 0.041 (\textbf{99$\times$}) \\
   peak memory (GB) & 18.275 & 2.612 (\textbf{7$\times$})\\
       \hline
    \end{tabular} \normalsize
    \vspace{-10pt}
\end{wraptable}

\paragraph{Overall Performance of CoMERA.} Table \ref{tab:train-DLRM} shows the testing accuracy, testing loss (measured as normalized CE), memory costs, and model sizes of CoMERA and uncompressed training.  CoMERA achieves similar accuracy as the uncompressed training, while CoMERA compresses the whole model by $99\times$ and saves $7\times $ peak memory cost (with consideration of the data and backend overhead) in the training process. The reduction of model size and memory cost mainly comes from the compact TTM tensor representation of large embedding tables. Standard uncompressed training is faster than CoMERA since DLRM is an embedding-intensive model, and the computation in the embedding table is look-up rather than matrix multiplications. However, CoMERA uses much less memory, saving 6.9X, 4.8X, and 3.1X memory for batch sizes 10000, 2000, and 4000, respectively.
Furthermore, CoMERA has a similar convergence curve and needs fewer iterations than standard training for DLRM, as shown in Figure \ref{fig:DLRM_convergence}.

% \begin{table}[t]
%     \centering
%     \caption{Speed up of optimized embedding on DLRM training. \zz{This is the look-up time, not the training time.}}
%     \label{tab:emb-DLRM}
%     \begin{tabular}{|c|c|c|c|c|}
%     \hline
%        \multirow{2}{*}{batch} &  \multicolumn{2}{c|}{time (s)} &  \multicolumn{2}{c|}{memory (MB)} \\
%        \cline{2-5}
%           & new & old & new & old \\
%         \hline
%        10000 & 807 & 1344 & 2612 & 9259 \\
%        20000 & 794 & 1182 & 3947 & 18385 \\
%        40000 & 791 & N/A & 6629 & N/A \\
%        \hline
%     \end{tabular}
% \end{table}

% \begin{minipage}{\textwidth}
%   \begin{minipage}[b]{0.79\textwidth}
%         \centering
%          \includegraphics[width=\textwidth]{figures/CoMERA_time_memory.png}
%          \captionof{figure}{Performance of optimized CoMERA on training DLRM.}
%   \end{minipage}
%   \begin{minipage}[b]{0.19\textwidth}
%         \centering
%     \captionof{table}{Training results on the DLRM model with a batch size $10,000$.}
%     \label{tab:train-DLRM}
%     \begin{tabular}{|c|c|c|}
%     \hline
%     & uncompressed & CoMERA \\
%     \hline
%    accuracy & 78.68\% &  78.76\% \\
%     normalized CE& 0.793 & 0.792 \\
%    model size (GB) & 4.081 & 0.041 (\textbf{99$\times$}) \\
%    peak memory (GB) & 18.275 & 2.612 (\textbf{7$\times$})\\
%        \hline
%     \end{tabular}
%   \end{minipage}
% \end{minipage}

% \begin{table}[t]
%     \centering
%     \caption{Training results on the DLRM model with a batch size $10,000$. \zz{You may organize the table with 3 columns and 5 rows, and combine it with Fig. 5 (i.e., put this table at the rightmost of Fig. 5).}}
%     \label{tab:train-DLRM}
%     \begin{tabular}{|c|c|c|c|c|}
%     \hline
%              & accuracy & normalized CE& model size (GB) & peak memory (GB) \\
%              \hline
%        uncompressed & 78.68\% & 0.793 &  4.081 & 18.275 \\
%       CoMERA & 78.76\% & 0.792 & 0.041 (\textbf{99$\times$}) & 2.612 (\textbf{7$\times$}) \\
%        \hline
%     \end{tabular}
% \end{table}

\vspace{-10pt}
\subsection{Comparison with GaLore and LTE}
\vspace{-5pt}
We compare our method with two recent low-rank compressed training frameworks: GaLore ~\cite{zhao2024galore} and LTE~\cite{huh2024training}. GaLore~\cite{zhao2024galore} reduces the memory cost by performing SVD compression on the gradient, and LTE represents the weights as the sum of parallel low-rank matrix factorizations. We evaluate their memory costs and training times per epoch on the six-encoder transformer model under different batch sizes. We do not compare the total training time because the training epochs of various methods are highly case-dependent. The CoMERA and GaLore achieve almost the same validation accuracy, 64\%, on the MNLI dataset. However, the LTE approach does not converge on the task using its default setting.

\vspace{-10pt}
\paragraph{Training Time Per Epoch.} We use rank $128$ for the low-rank gradients in GoLore, and rank $32$ and head number $16$ for the low-rank adapters in LTE. For a fair comparison, all methods are executed with CUDA graph to reduce the overhead of launching CUDA kernels. The runtimes per training epochs are reported in Figure \ref{fig:compare_galore_LTE}(a). For the LTE, we only report the results for batch sizes $32,64,128$ since it requires the batch size to be a multiple of the head number. Overall, our CoMERA is about $2\times$ faster than GaLore and $3\times$ faster than LTE for all batch sizes, because the forward and backward propagation using low-rank tensor-network contractions dramatically reduce the computing FLOPS. 

\vspace{-10pt}
\paragraph{Memory Cost.} Figure \ref{fig:compare_galore_LTE} (b) shows the memory cost of all three training methods. In the single-batch setting as used in~\cite{zhao2024galore}, our CoMERA method is $9\times$ more memory-efficient than Galore on the tested case (with consideration of data and back-end cost). As the batch size increases, the memory overhead caused by data and activation functions becomes more significant, leading to less memory reduction ratios. However, our proposed CoMERA still uses the least memory. %

We run the experiments on the RTX 3090 GPU. The work GaLore\cite{zhao2024galore} uses the RTX 4090 GPU for experiments, so we also compare them on the RTX 3090 GPU. The results are in Appendix \ref{appendix:Galore_LTE_4090}.

\vspace{-10pt}
\subsection{Preliminary LLM Pre-Training Results: Case Study on CodeBERT}
\begin{figure}
    \centering
    \vspace{-15pt}
    \includegraphics[width=\linewidth]{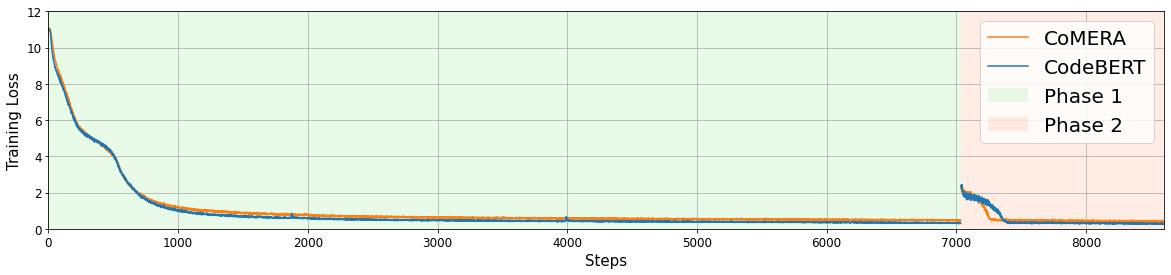}
    \caption{Pre-training loss curves of CodeBERT and CoMERA.}
    \vspace{-15pt}
    \label{fig:codebert-comera-loss}
\end{figure}

\vspace{-5pt}
To show the benefit of CoMERA in pre-training (domain-specific) LLMs, we follow the setup from CodeBERT \cite{feng2020codebert} to pre-train a BERT-like model for code generation. The pre-training dataset is the CodeSearchNet \cite{husain2019codesearchnet}, a collection of 2M (comment, code) pairs and 6M pure code sequences from open-source libraries with 6 types of programming languages. We pre-train $\text{CodeBERT}_{\text{LARGE}}$ ($357$M) and its CoMERA ($84$M) variant using the masked language modeling (MLM) objective and compare their training loss in Figure \ref{fig:codebert-comera-loss}. We achieve up to $12.72\times$ compression on tensorized layers and $4.23\times$ overall compression with final loss of $0.40$ vs $0.28$. There is a small gap between the final losses. However, this does not necessarily imply performance degradation on downstream tasks, based on our observation on $\text{BERT}_{\text{LARGE}}$\cite{devlin2019bert} shown in Appendix \ref{appendix:BERT_large}. Furthermore, CoMERA is $2.3\times$ and $1.9\times$ faster than standard pre-training in Phase 1 and Phase 2 respectively, when evaluated on the Nvidia RTX 3090 GPU. Our current CoMERA implementation is still slower than standard pre-training on massive GPUs, since no performance optimization has been done on HPC.

% \vspace{-10pt}
% \paragraph{CoMERA on Original $\text{BERT}_{\text{LARGE}}$.} Our results on CodeBERT is rather preliminary as only pre-training loss is available. For the original $\text{BERT}_{\text{LARGE}}$ (336M) and its CoMERA (125M) variant that we trained by using Wikipedia (2500M words), we achieve up to 6.36x compression on tensorized layers and 2.69x overall compression, with final loss of 1.45 vs 1.26. On downstream tasks, CoMERA ourperforms $\text{BERT}_{\text{LARGE}}$ on SST-2 (accuracy: 92.10\% vs 91.74\%) and MRPC (accuracy: 86.82\% vs 86.00\%), underperforms $\text{BERT}_{\text{LARGE}}$ on SQuAD (f1: 88.76\% vs 90.68\%).

\vspace{-10pt}
\section{Conclusions and Future work} \label{sc:conclusion}
\vspace{-5pt}
This work has presented CoMERA framework to reduce the memory and computing time of training AI models. We have investigated rank-adaptive training via multi-objective optimization to meet specific model sizes while maintaining model performance. We have achieved real training speedup on GPU via three optimizations: optimizing the tensorized embedding tables, optimizing the contraction path in tensorized forward and backward propagation, and optimizing the GPU latency via CUDAGraph. The experiments on a transformer model demonstrated that CoMERA can achieve $2-3\times$ speedup per training epoch. The model sizes of the transformer and a DLRM model have been reduced by $43\times$ to $99\times$ in the training process, leading to significant peak memory reduction (e.g., $7\times$ total reduction in large-batch training of DLRM on a single GPU). Our method has also outperformed the latest GaLore and LTE frameworks in both memory and runtime efficiency. More importantly, our method has demonstrated significant speedup and model compression in pre-training CodeBERT, a domain-specific LLM for automatic code generation. We have also observed further speedup by combining CoMERA with mixed-precision computation. The discussions and some preliminary results are in Appendix \ref{appendix:mixed-precision}. 

Unlike large-size matrix operations, the small low-rank tensor operations used in CoMERA are not yet well-supported by existing GPU kernels. The performance of CoMERA can be further boosted significantly after a comprehensive GPU and HPC optimization. The existing optimizers, e.g. Adam, are well-studied for uncompressed training. However, CoMERA has a very different optimization landscape due to the tensorized structure. Therefore, it is also worth studying the optimization algorithms specifically for CoMERA in the future.

\vspace{-10pt}
\section{Acknowledgement}
\vspace{-10pt}
The pre-training task used resources of the National Energy Research Scientific Computing Center, a DOE Office of Science User Facility supported by the Office of Science of the U.S. Department of Energy
under Contract No. DE-AC02-05CH11231 using NERSC award ASCR-ERCAP0030039.

\bibliographystyle{plain}
\bibliography{my_ref,zhang}

\appendix

\newpage
\section{Supplementary Material}

% \subsection{Some figures for Tensor Contractions and Decompositions in Section \ref{sc:preliminary}}
% Fig. \ref{fig:tensor}(a) illustrates tensors of various orders with corresponding tensor network representations and Figure~\ref{fig:tensor}(b) illustrates some tensor contractions. 
% \begin{figure}[t]
%     \centering
%     \includegraphics[width=\textwidth]{figures/tensor_and_contractions.png}
%     \caption{(a) Tensors. (b) Tensor contractions.}
%     \label{fig:tensor}
% \end{figure}

% The TT and TTM decomposition in the tensor network representation is illustrated in Figure \ref{fig:TT_network}(a) and \ref{fig:TT_network}(b) respectively.
% \begin{figure}[t]
%     \centering
%     \includegraphics[width=0.8\textwidth]{figures/TT_TTM_contractions_new.png}
%     \caption{Tensor networks for (a) tensor-train and (b) tensor-train-matrix decompositions.}
%     \label{fig:TT_network}
%     \vspace{-10pt}
% \end{figure}

% \subsection{Figure for TTM embedding in Section \ref{sc:embedding}}
% \label{appendix:embedding}
% The optimized TTM embedding table lookup in tensor network representation is shown in Figure \ref{fig:TTM-emb}.
% \begin{figure}
%          \centering
%          \includegraphics[width=2.8in]{figures/TTM_emb_new.png}
%     \caption{Optimized TTM embedding table lookup.}
%          \label{fig:TTM-emb}
% \end{figure}

\subsection{Comparison with existing works}
\label{appendix:comparison_existing}
SVDinsTN~\cite{zheng2024svdinstn} uses sparse diagonal matrices and the $\ell_1$ regularization to control tensor ranks for a compact tensor structure of a given tensor. In contrast, our work compresses weights during end-to-end training without any prior information on the tensor (i.e., model parameters). Both works use sparse diagonal matrices and $\ell_1$ terms to control tensor ranks. Using diagonal matrices to control the ranks of matrices is very common, like SVD. The $\ell_1$ norm is also widely used to induce sparsity in various models, like compressed sensing and Lasso regression. It is natural to combine these two techniques to control tensor ranks, regardless of tensor formats. Moreover, our work formulates the problem as a more generic multi-objective problem and uses a two-stage algorithm to solve it. The formulation in SVDinsTN is similar to linear scalarization approach in our early stage. Our work further uses the achievement scalarization in the late stage to find a model close to our preferred model performance and size.

HEAT \cite{gu2022heat} also considers contraction optimization for post-training model compression of trained models. In contrast, CoMERA considers end-to-end tensor-compressed training, where no model parameters are known prior to training. In addition, HEAT only discusses the single path optimization for forward propagation in CP format. We have optimized $d+2$ contraction paths {\bf jointly} in both forward- and back- propagation in TT format. Since these contractions can be coupled, we have also minimized the overall computation costs by reusing intermediate results.

\subsection{Proof of Proposition \ref{prop:Pareto}}
\label{sec:proof:prop_Pareto}

\begin{proof}
    The objective function in \eqref{eq:pretrain-rank} is bounded below by $0$. Hence, the problem \eqref{eq:pretrain-rank} has a finite infimum value $f^*$. Let $\{\ten{G}^k,\mat{D}^k\}_{k=1}^{\infty}$ be a sequence such that $\lim_{k\to\infty} L(\ten{G}^k,\mat{D}^k)+\gamma \hat{S}(\mat{D}^k)+\beta \|\ten{G}^k\|^2 = f^*$. The sequence must be bounded because of the $l_1$ regularization of $\mat{D}$ and the $\ell_2$ regularization of $\ten{G}$. As a result, the sequence has a cluster point $(\ten{G}^*,\mat{D}^*)$ which is a minimizer of the \eqref{eq:pretrain-rank}. Let $C:=\|\ten{G}^*\|^2$. The relaxation \eqref{eq:pretrain-rank} is equivalent to the constrained optimization problem. 
\begin{eqnarray} \label{eq:constrained-pretrain-rank}
    \min_{\ten{G},\mat{D}} & L(\ten{G},\mat{D})+\gamma \hat{S}(\mat{D}) \\ \nonumber
    \text{s.t.} & \|\ten{G}\|^2 \le C.
\end{eqnarray}
It implies that the solution to the training problem \eqref{eq:pretrain-rank} is a Pareto point of the multi-objective optimization problem $\min_{\ten{G},\mat{D}} (L(\ten{G},\mat{D}), \hat{S}(\mat{D}))$. 
\end{proof}

\subsection{Algorithm for Late Stage Optimization in Section \ref{sc:training methods}}
\label{appendix:late stage}
The algorithm for the late stage optimization in Section \ref{sc:training methods} is summarized in Algorithm \ref{alg:scalar}.

\begin{algorithm}[h]
\caption{Solve relaxed scalarization problem \eqref{eq:scalarization relax}} 
\label{alg:scalar}
\hspace*{\algorithmicindent} \textbf{Input:} Initializations $\ten{G}_0,\mat{D}_0$, constants $L_0,S_0,w_1,w_2,\rho,\beta$, and an optimization algorithm $\mathcal{O}$.\\
\hspace*{\algorithmicindent} \textbf{Output:} Tensor cores $\ten{G}_T$ and rank-control parameters $\mat{D}_T$.
\begin{algorithmic}[0]
\For{$t=0,\ldots,T-1$}
\If{$w_1 (L(\ten{G}_t,\mat{D}_t)-L_0)\ge w_2 ({S}(\mat{D}_t) - S_0)$}
\State The optimization algorithm $\mathcal{O}$ runs one step on the problem \eqref{eq:1>2}.
\Else 
\State The optimization algorithm $\mathcal{O}$ runs one step on the problem \eqref{eq:1<2}. 
\EndIf
\EndFor
\end{algorithmic}
\end{algorithm}

\subsection{Proof of Proposition \ref{prop:forward}}
\label{appendix:path proof}
% \begin{proposition} \label{prop:forward formal}
%     Suppose that the TT ranks satisfy $1=r_0< r_1 \le \cdots \le r_d \ge r_{d-1} > \cdots \ge r_{2d}=1$. There exists some $M>0$ such that for all batch sizes $b>M$, the following holds. In the contraction path with the least number of flops for the forward-propagation \eqref{eq:forward}, let $\ten{V}_1,\ldots,\ten{V}_k$ be the tensors after contractions without $b$ ordered by their occurrences in the remaining path, then each $\ten{V}_i$ has dimension $r_{j_i}\times n_{j_i+1} \times \cdots \times n_{j_{i+1}}\times r_{j_{i+1}}$ for $0= j_1<\cdots <j_k<j_{k+1}=2d$.
% \end{proposition}

\begin{proof}
    For convenience, let $V_i$ be a string of characters to specify the dimension of $\ten{V}_i$, $C_i$ be the set of tensor cores used to obtain $\ten{V}_i$, and $\ten{X}_i$ be the tensor by contracting $\ten{X}$ with $\ten{V}_1,\ldots.\ten{V}_i$, denoted by the string $X_i$.

    We first show that $\ten{V}_1$ must be in the proposed format. Suppose otherwise for contradiction. Let $\ten{G}_i$ be the first tensor core used to obtain $\ten{V}_1$. If $i=1$, then we write $V_1={V}_1^1{V}_1^2$ where the tensor $\ten{V}_1^1$ corresponding to $V_1^1$ is obtained by contractions of longest consecutive tensor cores containing $\ten{G}_i$ in the set $C_1$. Let $Z_1 = {V}_i^1 \cap \tilde{X}$, $Z_2 = {V}_i^2 \cap \tilde{X}$. The number of flops for the contraction between $\tten{X}$ and $\ten{V}_1$ is $\frac{\pi(X)\pi(V_1^1)\pi(V_1^2)}{\pi(Z_1)\pi(Z_2)}$. If we first contract $\tten{X}$ with $\ten{V}_1^1$ and then contract the obtained tensor with $\ten{V}_1^2$, the number of flops is $\frac{\pi(X)\pi(V_1^1)}{\pi(Z_1)}+\frac{\pi(X)\pi(V_1^1)\pi(V_1^2)}{\pi(Z_1)^2\pi(Z_2)}$ which is less than $\frac{\pi(X)\pi(V_1^1)\pi(V_1^2)}{\pi(Z_1)\pi(Z_2)}$ since $Z_2 \neq V_1^2$. It contradicts our assumption that this is the optimal path. If $d\ge i>1$, let $\ten{S}$ be the tensor generated by the longest consecutive tensor cores containing $\ten{G}_{i-1}$ and used in the optimal path. A better path is to first contract $\ten{V}_1$ with $\ten{S}$ to obtain $\ten{W}$, then contract $\ten{W}$ with $\ten{X}$ and all other unused parts in the optimal path. It is better because the number of flops for contracting $\ten{W}$ and $\ten{X}$ is no greater than that for contracting $\ten{V}_1$ and $\ten{X}$ and the new path reduces the number of flops in the remaining contractions. The reduction in the remaining contractions is more than the potential flop increase in obtaining $\ten{W}$ when the batch size $b$ is big enough. Finally, we consider the case that $i>d$. Let $\ten{G}_j$ be the first tensor core used to obtain $\ten{V}_2$. If $j>d$, then first contracting $\ten{V}_1$ and $\ten{V}_2$ and then all other parts is a better choice.  Otherwise, if $j\le d$, let $\ten{V}_2^1$ be the tensor contracted by the consecutive tensors containing $\ten{G}_j$ in $C_2$. The tensor $\ten{V}$ can be represented by the contraction of $\ten{V}_2^1$ and another tensor $\ten{V}_2^2$ generated by the remaining tensor cores in $C_2$. In this scenario, we can contract $\ten{V}_2^1$ with $\ten{V}_1$ to get $\ten{S}$, then $\ten{S}$ with $\ten{X}$ to get $\ten{W}$, then $\ten{V}_2^2$ with $\ten{W}$, and finally the obtained tensor with $\ten{V}_3,\ldots,\ten{V}_k$. It is not hard to verify contracting $\ten{V}_2^1$ and $\ten{V}_1$, $\ten{S}$ and $\ten{S}$ and $\ten{X}$, and $\ten{W}$ and $\ten{V}_2^2$ uses less flops than directly contracting $\ten{X}$ with $\ten{V}_1$ and $\ten{V}_2$ directly when the batch size $b$ is large. Summarizing everything above, we can conclude that $\ten{V}_1$ must be in the proposed format. 

    The contraction of $\ten{X}_i$ and $\ten{V}_{i+1},\ldots,\ten{V}_{k}$ has the similar structure to the contraction of $\ten{X}$ and $\ten{V}_1,\ldots,\ten{V}_{k}$. By applying the same proof, we conclude that the tensors $\ten{V}_i$'s must be in the format stated in the proposition and we will contract the input tensor $\ten{X}$ with the tensors $ten{V}_1,\ldots,\ten{V}_k$ in the sequential order.
\end{proof}

\subsection{Algorithm for Contraction Path in Section \ref{subsec:path}}
\label{sec:contract-algorithm}
The empirical near-optimal contraction path for tensor-compressed training is shown in Algorithm \ref{alg:path}. 

\begin{figure}[h]
     \centering
     \begin{subfigure}[b]{0.8\textwidth}
         \centering
         \includegraphics[width=\textwidth]{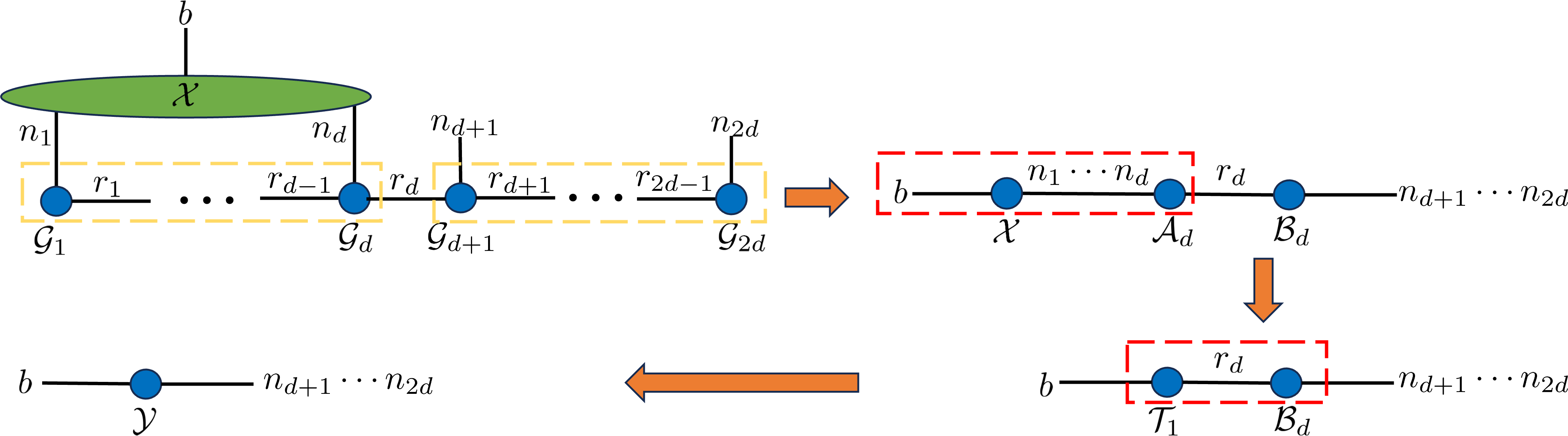}
         \caption{TT-vector forward-propagation}
     \end{subfigure}
     \hfill
     \begin{subfigure}[b]{0.8\textwidth}
         \centering
         \includegraphics[width=\textwidth]{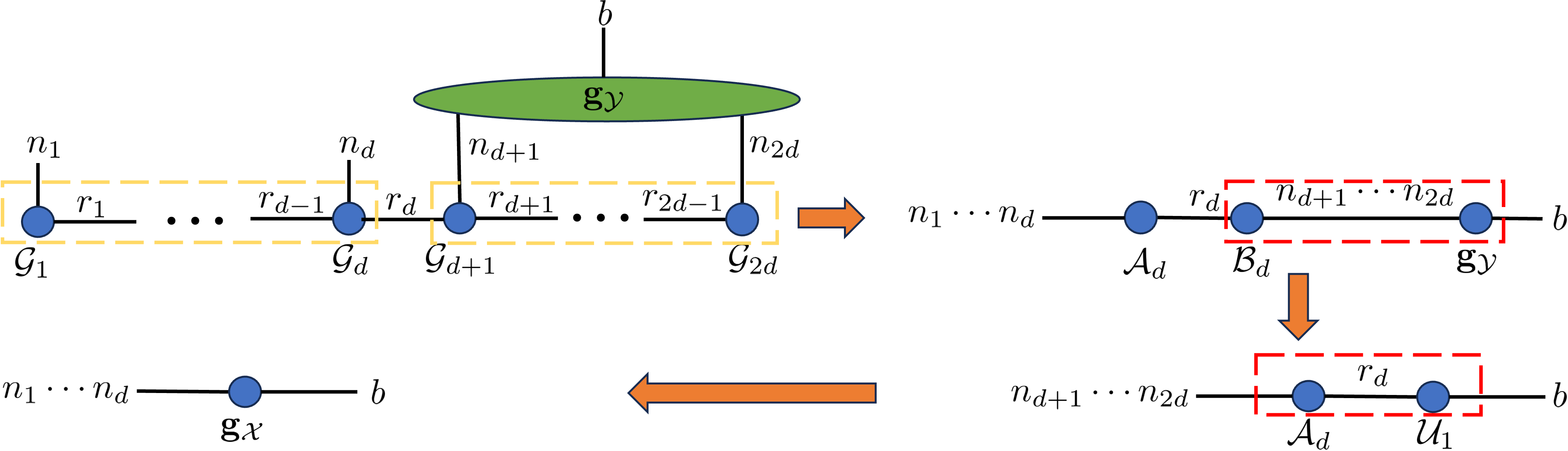}
         \caption{TT-vector back-propagation: compute $\mat{g}_{\mat{X}}$}
     \end{subfigure}
     \begin{subfigure}[b]{0.8\textwidth}
         \centering
         \includegraphics[width=\textwidth]{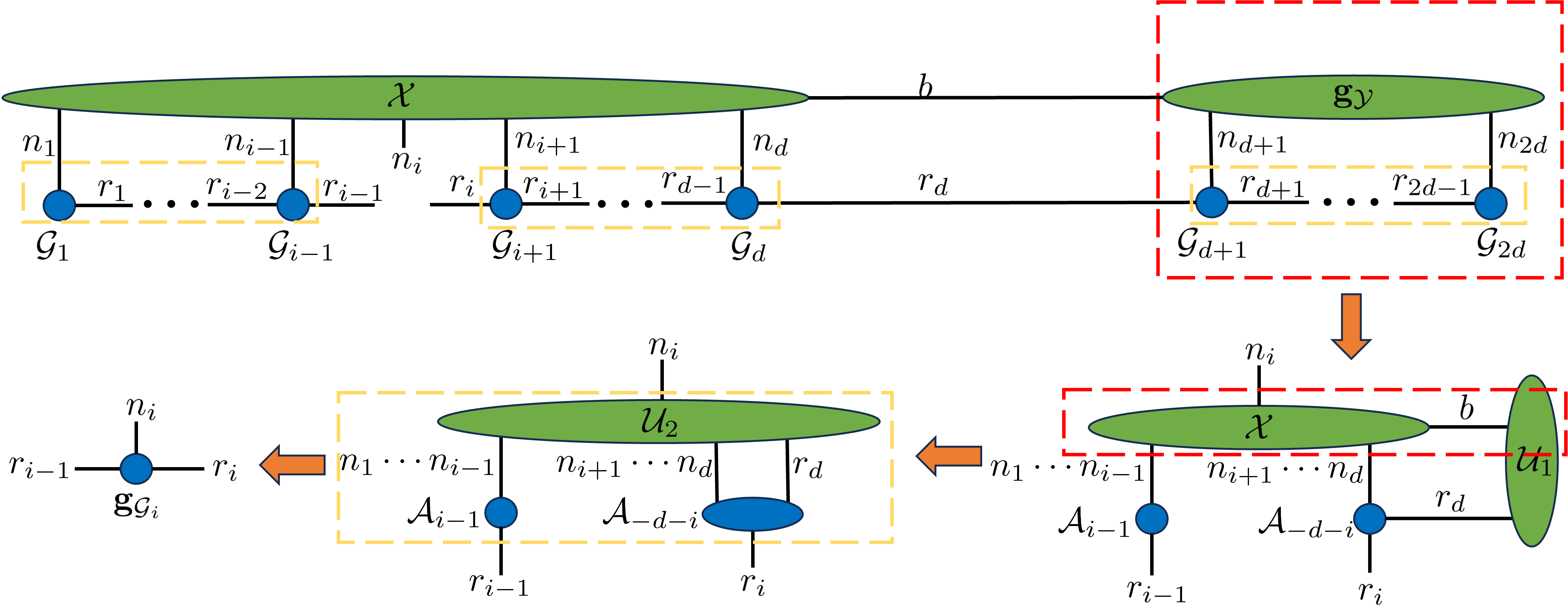}
         \caption{TT-vector back-propagation: compute $\mat{g}_{\mat{G}_i},1\le i\le d$}
     \end{subfigure}
     \hfill
     \begin{subfigure}[b]{0.8\textwidth}
         \centering
         \includegraphics[width=\textwidth]{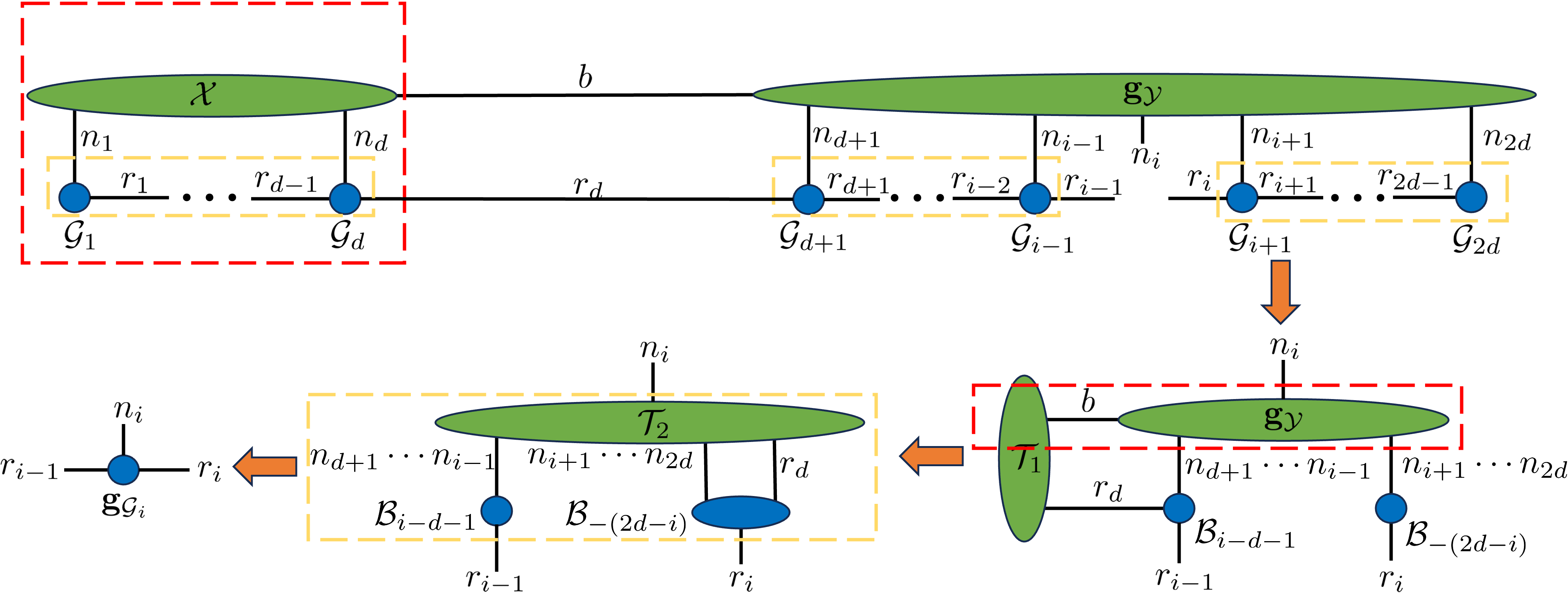}
         \caption{TT-vector back-propagation: compute $\mat{g}_{\mat{G}_i},d+1 \le i \le 2d$}
     \end{subfigure}
     \caption{Tensor diagrams for contraction paths of TT forward- and back- propagation.}
     \label{fig:TT_path}
\end{figure}

Figure \ref{fig:TT_path} presents the tensor diagrams for contraction paths of TT forward- and back- propagation as discussed in Section \ref{subsec:path}.

\begin{algorithm}[h]
\caption{Empirical path for tensor-compressed forward- and back- propagation} 
\label{alg:path}
 \textbf{Forward Input:} Tensor cores $\ten{G}_1,\ldots,\ten{G}_{2d}$ and input matrix $\mat{X}$.\\
\textbf{Forward Output:} Output matrix $\mat{Y}$, and intermediate results.
\begin{algorithmic}[1]
\State Reshape the matrix $\mat{X}$ to the tensor $\ten{X}$.
\State Compute $\ten{A}_d,\ten{A}_{-d},\ten{B}_d,\ten{B}_{-d}$ in the sequential order and store intermediate results of $\{\ten{A}_i,\ten{A}_{-i},\ten{B}_i,\ten{B}_{-i}\}_{i=1}^d$ for back-propagation.
\State Compute $\ten{T}_1$ as in \eqref{eq:path-T1} to store it for back-propagation.
\State Compute $\ten{Y}$ as in \eqref{eq:path-Y} and reshape it to the appropriate matrix $\mat{Y}$.
\end{algorithmic}
 \textbf{Backward Input:} Inputs of \textbf{Forward}, stored results from \textbf{Forward}, and output gradient $\mat{g}_{\mat{Y}}$.\\
\textbf{Backward Output:} Gradients $\mat{g}_{\mat{X}},\mat{g}_{\ten{G}_1},\ldots,\mat{g}_{\ten{G}_{2d}}$.
\begin{algorithmic}[1]
\State Reshape the gradient $\mat{g}_{\mat{Y}}$ to the tensor $\mat{g}_{\ten{Y}}$.
\State Compute $\ten{U}_1$ and $\mat{g}_{\ten{X}}$ as in \eqref{eq:path-U1}, \eqref{eq:path-gX} and store $\ten{U}_1$ for future use.
\State Compute $\mat{g}_{\ten{G}_i}$ for $i\ge d+1$ as in \eqref{eq:path-T2}, \eqref{eq:path-gGi>d} using stored tensors.
\State Compute $\mat{g}_{\ten{G}_i}$ for $i\le d$ as in \eqref{eq:path-T1}, \eqref{eq:path-gGi<d} using stored tensors.
\end{algorithmic}
\end{algorithm}

\label{appendix:TT_path}

\subsection{Small Batch Case for Contraction Path in Section \ref{subsec:path}}
\label{appendix:path}

\paragraph{Small batch case.} The empirical contraction path in Algorithm \ref{alg:path} eliminates the batch size dimension $b$ early, so it is nearly optimal when the batch size is large. 
We may search for a better path using a greedy search algorithm to minimize the total operations. In each iteration, we prioritize the pairs that output the smallest tensors. Such a choice can quickly eliminate large intermediate dimensions to reduce the total number of operations. When the batch size is large, the searched path is almost identical to the empirical path in Algorithm \ref{alg:path} which eliminates the batch size dimension $b$ early. The searched path may differ from Algorithm \ref{alg:path} for small batch sizes, but their execution times on GPU are almost the same. This is because the tensor contractions for smaller batch sizes have a minor impact on the GPU running times. Consequently, despite certain tensor contractions in the empirical path being larger than those in the optimal path, the actual GPU execution times between them exhibit only negligible differences. Therefore, the empirical contraction path in Algorithm \ref{alg:path} is adopted for all batch sizes in CoMERA.

\subsection{Compression Settings for the Experiment in Section \ref{sc:6encoder}}
\label{appendix:settings}
The compression settings for the experiment in Section \ref{sc:6encoder} are shown in Table \ref{tab:tensor BERT}.

\begin{table}[t]
    \centering
    \footnotesize	
    \caption{Tensorized setting for the Transformer model in CoMERA.}
    \label{tab:tensor BERT}
    \begin{tabular}{|c|c|c|c|c|}
    \hline
     & format & linear shape &tensor shape & rank \\
     \hline
       embedding  & TTM & (30527,768) & (64,80,80,60) & 30\\
        attention & TT & (768,768) & (12,8,8,8,8,12) & 30\\
        feed-forward & TT & (768,3072) & (12,8,8,12,16,16) & 30\\
        \hline
    \end{tabular}
\end{table}

\subsection{Per Epoch Training Time of CoMERA on MNLI for Various Compression Ratios}
\label{appendix:training_MNLI_ratios}
\begin{figure}
    \centering
    \includegraphics[width=0.8\linewidth]{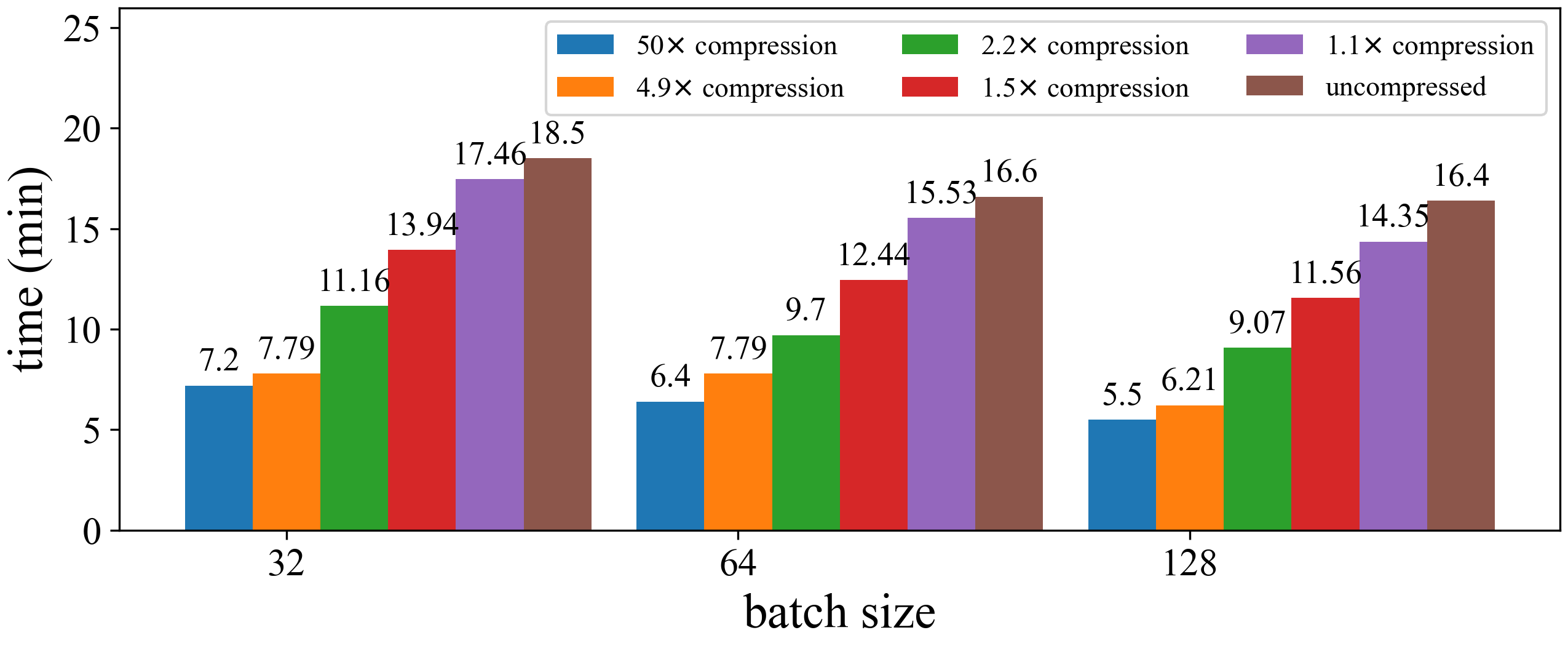}
    \caption{Per epoch training time of CoMERA on MNLI for various compression ratios.}
    \label{fig:training_MNLI_ratios}
\end{figure}

Table \ref{fig:training_MNLI_ratios} shows the per-epoch training time of CoMERA on MNLI dataset for different compression ratios. The acceleration is more obvious for larger compression ratios. When the compression ratio is greater than 1, CoMERA always has speedup. When the compression ratio approaches 1, the time of CoMERA approaches that of uncompressed training.

% \begin{table}[t]
%     \centering
%     \caption{Per epoch training time of CoMERA on MNLI for various compression ratios.}
%     \label{tab:training_MNLI_ratios}
%     \begin{tabular}{|c|c|c|c|c|c|c|c|}
%     \hline
%        \multicolumn{2}{|c|}{} & rank 30 & rank 120 & rank 240 & rank 360 & rank 480 & uncompressed\\
%        \hline
%        \multicolumn{2}{|c|}{compression ratio} & 50$\times$ &4.9$\times$ &2.2$\times$ &1.5$\times$ &1.1$\times$ & N/A \\
%        \hline
%        \multirow{3}{*}{time (min)} & batch 32 & 7.2&	7.79	&11.16	&13.94	&17.46&	18.5\\
%        \cline{2-8}
%         & batch 64 & 6.4	&6.73	&9.7&	12.44&	15.53	&16.6 \\
%        \cline{2-8}
%         & batch 128 & 5.5&	6.21	&9.07	&11.56	&14.35&	16.4
%  \\
%        \cline{1-8}
%     \end{tabular}
% \end{table}

\subsection{Comparison with GaLore and LTE on a single RTX 4090 GPU}
\label{appendix:Galore_LTE_4090}
\begin{figure}[t]
     \centering
     \begin{subfigure}[b]{0.48\textwidth}
         \centering
         \includegraphics[width=\textwidth]{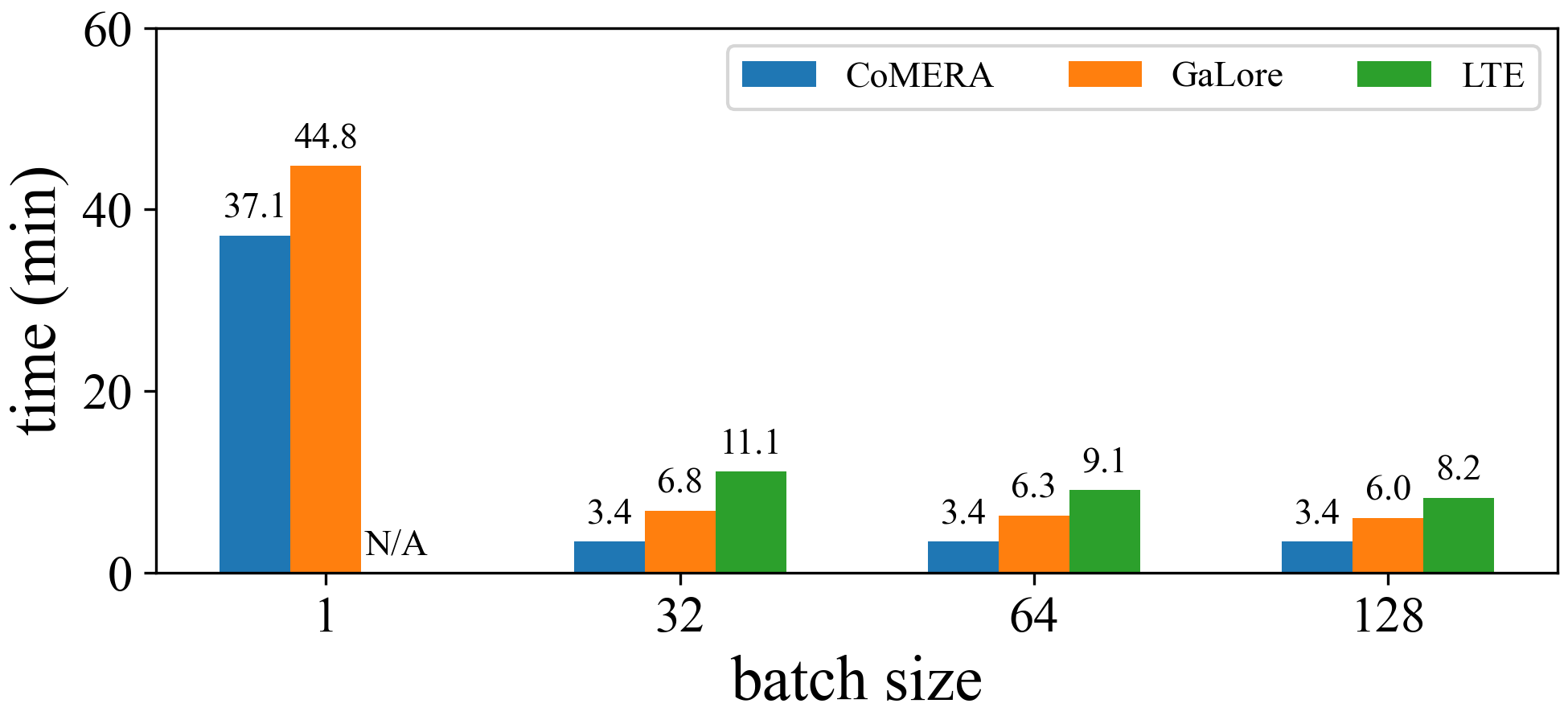}
         \caption{Training time per epoch.}
     \end{subfigure}
     \hfill
     \begin{subfigure}[b]{0.48\textwidth}
         \centering
         \includegraphics[width=\textwidth]{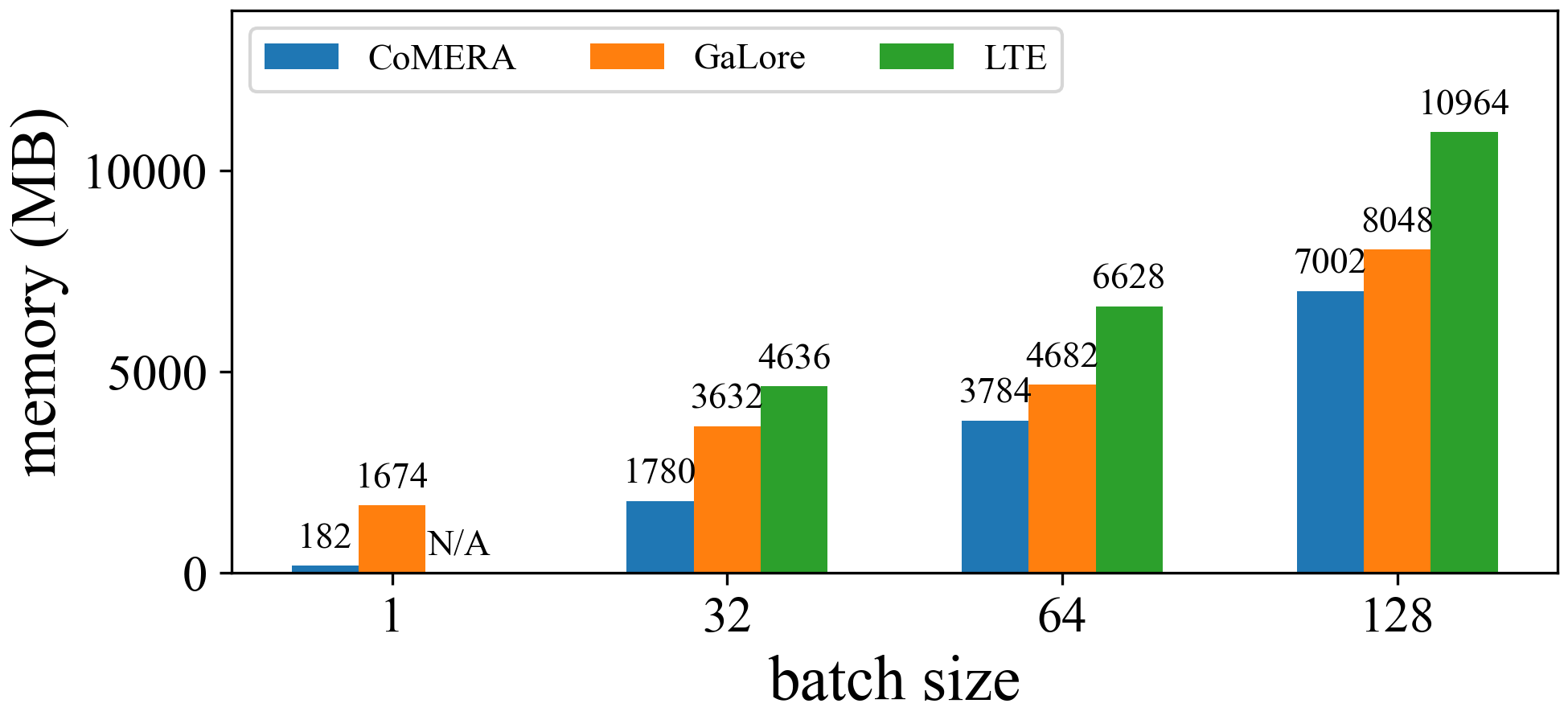}
         \caption{Peak memory consumption.}
     \end{subfigure}
     \caption{Training time and total memory cost of CoMERA, GaLore~\cite{zhao2024galore} and LTE~\cite{huh2024training} on a six-encoder transformer with varying batch sizes. The experiment is done on Nvidia RTX 4090 GPU.}
     \label{fig:compare_galore_LTE_4090}
     \vspace{-15pt}
\end{figure}

Since GaLore\cite{zhao2024galore} uses a single RTX 4090 GPU for experiments in the original paper, we also run the experiments on the RTX 3090 GPU and compare the results. Figure \ref{fig:compare_galore_LTE_4090} presents the training time and peak memory consumption. Compared to RTX 3090, the training on RTX 4090 uses similar memory and takes less training time, and CoMERA is still the fastest method and consumes the least memory among all three techniques. The memory savings are almost the same as the results reported in Figure 1 in our paper. The speed-up factors are almost identical for batch sizes 32, 64, and 128. For batch size 1, our method is $1.2\times$ faster and $1.7\times$ faster than GaLore on RTX 3090, respectively. The difference is that RTX 4090 GPU significantly accelerates matrix multiplications of batch size 1, while it does not accelerate that much for smaller tensor contractions. We find that $r=30$ matrix multiplication on RTX 3090 has a similar speedup for both batch sizes, whereas the same multiplication on RTX 4090 only has speedup for batch 32 and does not have any speedup for batch 1. We would like to note that it might be caused by that different GPU platforms have different backend overhead, which can become more dominant as computation decreases to batch=1. We will continue optimizing GPU-level kernels to accelerate small tensor contractions and expect to see a similar speedup.

\subsection{CoMERA Pretraining Result on $\text{BERT}_{\text{LARGE}}$}
\label{appendix:BERT_large}
\paragraph{CoMERA on Original $\text{BERT}_{\text{LARGE}}$.} Our results on CodeBERT is rather preliminary as only pre-training loss is available. For the original $\text{BERT}_{\text{LARGE}}$\cite{devlin2019bert} (336M) and its CoMERA (125M) variant that we trained by using Wikipedia (2500M words), we achieve up to $6.36\times$ compression on tensorized layers and $2.69\times$ overall compression, with final loss of 1.45 vs 1.26. On downstream tasks, CoMERA ourperforms $\text{BERT}_{\text{LARGE}}$ on SST-2 (accuracy: 92.10\% vs 91.74\%) and MRPC (accuracy: 86.82\% vs 86.00\%), underperforms $\text{BERT}_{\text{LARGE}}$ on SQuAD (f1: 88.76\% vs 90.68\%).

\subsection{Discussion: Mixed-Precision CoMERA}
\label{appendix:mixed-precision}
Modern GPUs offer low-precision computation to speed up the training and inference. It is natural to combine low-rank tensor compression and quantization to achieve the best training efficiency. However, CoMERA involves many small-size low-rank tensor contractions, and a naive low-precision implementation may even slow down the training due to the overhead caused by precision conversions. 

To resolve the above issue, we implement mixed-precision computation in CoMERA based on one simple observation: large-size contractions enjoy much more benefits of low-precision computation than small-size ones. This is because the overhead caused by precision conversions can dominate the runtime in small-size contractions. In large-batch tensor-compressed training, small- and large-size tensor contractions can be distinguished by whether the batch size dimension $b$ is involved. In general, a contraction with the batch $b$ is regarded as large and is computed in a low precision. Otherwise, it is regarded small and is computed in full-precision. The actual mixed-precision algorithm depends on the contraction path used in the forward- and back- propagations of CoMERA. %The path in Algorithm \ref{alg:path} is near-optimal for large batch sizes. Algorithm \ref{alg:mixed} illustrates the mixed-precision training algorithm on the path. In Figure \ref{fig:TT_forward_back} that shows the tensor-vector forward and backward passes, all contractions circled in red are computed in low-precision while all contractions circled in yellow are computed in a high precision.

\vspace{-5pt}
\begin{wrapfigure}{r}{3in}
         \centering
         \vspace{-10pt}
         \includegraphics[width=3in]{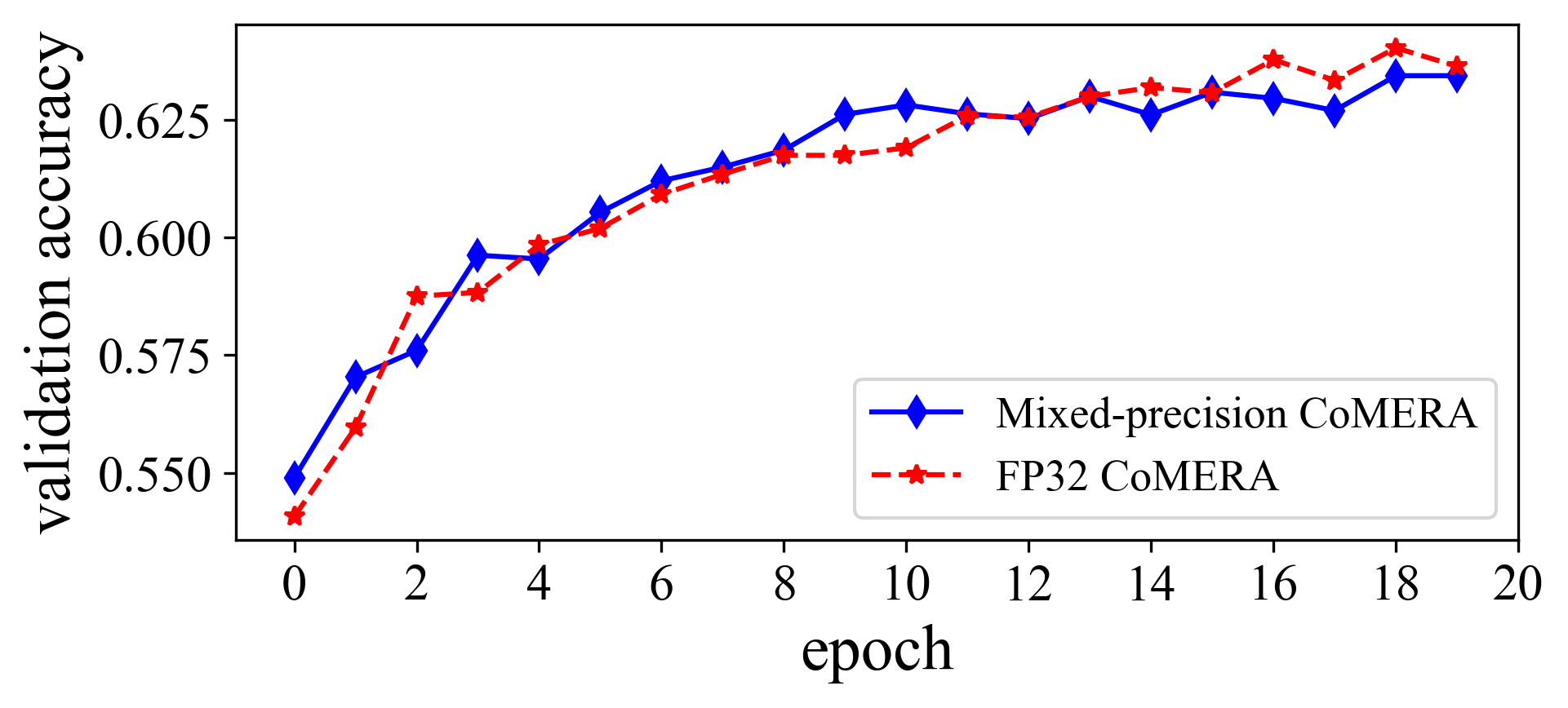}
     \caption{Convergence of mixed-precision CoMERA on the six-encoder transformer.}
         \label{fig:MNLI-mixed}
\end{wrapfigure}
\paragraph{Runtime.} We evaluate the mixed-precision forward and backward propagations of CoMERA in a FP8 precision on the NVIDIA L4 GPU. We consider a single linear layer. The shapes $(1024,1024)$ and $(1024,4096)$ are converted to the TT shapes $(16,8,8,8,8,16)$ and $(16,8,8,16,16,16)$ respectively, and the ranks are both $32$. The total execution time for $1000$ forward and backward propagations are shown in Table \ref{tab:LP-time}. The FP8 tensor-compressed linear layer has about $3\times$ speed-up compared to the FP8 vanilla linear layer when the batch size and layer size are large. When the batch size is small, the FP8 vanilla linear layer is even faster. This is because the tensor-compressed linear layer consists of a few sequential computations that are not well supported by current GPU kernels. We expect to see a more significant acceleration after optimizing the GPU kernels.

\begin{table}[t]
    \centering
    \caption{Speed-up of mixed-precision computation on tensor-compressed linear layers.}
    \label{tab:LP-time}
    \begin{tabular}{|c|c|c|c|c|}
    \hline
       \multirow{2}{*}{shape (b,m,n)}  & \multicolumn{2}{|c|}{tensor-vector} & \multicolumn{2}{|c|}{matrix-vector} \\
         \cline{2-5}
         & FP8-mix & FP32 & FP8 & FP32 \\
         \hline
        (10000,1024,1024) & 1.95 &  1.63 & 1.02 & 2.82 \\
        (20000,1024,1024) & 2.02 &  3.37 & 1.93 & 5.41 \\
        (40000,1024,1024) & 2.55 &  6.82 & 4.27 & 10.93 \\
        (10000,1024,4096) & 1.97 &  3.96 & 3.69 & 10.28 \\
        (20000,1024,4096) & 2.96 &  8.32 & 7.26 & 20.93 \\
        (40000,1024,4096) & 5.47 &  17.13 & 15.27 & 45.60 \\
        \hline
    \end{tabular}
\end{table}

\vspace{-5pt}
\begin{table}[h]
    \centering
    \caption{Training results of mixed-precision CoMERA on DLRM (batch size=10,000).}
    \label{tab:train-DLRM-mixed}
    \begin{tabular}{|c|c|c|}
    \hline
             & accuracy & normalized CE  \\
             \hline
       FP32 CoMERA & 78.76\% & 0.792 \\
       FP8/FP32 mixed-precision CoMERA & 78.88\% & 0.793 \\
       \hline
    \end{tabular}
\end{table}
\paragraph{Convergence.} We use the mixed-precision CoMERA to train the DLRM model and the six-encoder transformer. The result on DLRM is shown in Table \ref{tab:train-DLRM-mixed}. The convergence curve of the six-encoder transformer is shown in Figure \ref{fig:MNLI-mixed}. The experiments demonstrate that the accuracy of FP8 training is similar to FP32 training. However, we did not see much acceleration of using FP8 in the experiments. This is mainly because of \textcircled{1} the computation overhead of slow data type casting between FP32 and FP8; \textcircled{2} the sequential execution of small tensor contractions that are not well supported by current GPUs; \textcircled{3} the relatively small sizes of linear layers in the tested models. We will investigate these problems in the future, and are optimistic to see significant acceleration on larger models. %We are  by using mixed-precision precision on CoMERA.

\begin{comment}
\begin{figure}
    \centering
    \includegraphics[width=0.6\textwidth]{figures/MNLI_FP8.png}
    \caption{Convergence of mixed-precision CoMERA on MNLI dataset}
    \label{fig:MNLI-mixed}
\end{figure}
\end{comment}

% \input{Checklist}
% \section*{References}

% References follow the acknowledgments in the camera-ready paper. Use unnumbered first-level heading for
% the references. Any choice of citation style is acceptable as long as you are
% consistent. It is permissible to reduce the font size to \verb+small+ (9 point)
% when listing the references.
% Note that the Reference section does not count towards the page limit.
% \medskip

% {
% \small

% [1] Alexander, J.A.\ \& Mozer, M.C.\ (1995) Template-based algorithms for
% connectionist rule extraction. In G.\ Tesauro, D.S.\ Touretzky and T.K.\ Leen
% (eds.), {\it Advances in Neural Information Processing Systems 7},
% pp.\ 609--616. Cambridge, MA: MIT Press.

% [2] Bower, J.M.\ \& Beeman, D.\ (1995) {\it The Book of GENESIS: Exploring
%   Realistic Neural Models with the GEneral NEural SImulation System.}  New York:
% TELOS/Springer--Verlag.

% [3] Hasselmo, M.E., Schnell, E.\ \& Barkai, E.\ (1995) Dynamics of learning and
% recall at excitatory recurrent synapses and cholinergic modulation in rat
% hippocampal region CA3. {\it Journal of Neuroscience} {\bf 15}(7):5249-5262.
% }

%%%%%%%%%%%%%%%%%%%%%%%%%%%%%%%%%%%%%%%%%%%%%%%%%%%%%%%%%%%%

\end{document}